\documentclass{article}
\usepackage{graphicx} % Required for inserting images
\usepackage{amsmath}
\usepackage[left=2.5cm, right=2.5cm]{geometry}
\usepackage{color} 
\usepackage{amssymb}
\usepackage{accents}
\usepackage{ulem}
\usepackage{float}
\usepackage{caption}
\usepackage{hyperref}
\usepackage{algorithm}
\usepackage{algpseudocode}
\usepackage[english]{babel}
\usepackage{bbm}
\usepackage{chngcntr}
\usepackage{enumitem}
\usepackage{hyperref}
\usepackage{lmodern} % ordentliche Schriften
\usepackage{setspace,graphicx,tikz,tabularx}
\usepackage{booktabs}
\usepackage{dsfont}
\newcommand{\R}{{\mathbb{R}}}
\newcommand{\E}{{\mathbb{E}}}
\def\rd#1{{\color{black}#1}}

\DeclareMathOperator{\Tr}{Tr}

\newtheorem{theorem}{Theorem}[section]

\newtheorem{remark}{remark}[section]

\usepackage{fancyhdr}

\newcommand{\dps}[1]{\displaystyle{#1}}

\begin{document}
\author{Thomas \sc{Deschatre}\footnote{EDF R\&D \& FiME \sf \href{mailto:thomas.deschatre at edf.fr}{thomas.deschatre at edf.fr}}
\and
Xavier \sc{Warin}
\footnote{EDF R\&D \& FiME \sf \href{mailto:xavier.warin at edf.fr}{xavier.warin at edf.fr}}}
\title{Input Convex Kolmogorov Arnold Networks}
\maketitle

\begin{abstract}
This article presents an input convex neural network architecture using Kolmogorov-Arnold networks (ICKAN).
Two specific networks are presented: the first is based on a low-order, piecewise-linear representation of functions, and a universal approximation theorem is provided. The second is based on cubic splines, for which only numerical results support convergence.
We demonstrate   through simple tests that these networks perform competitively with classical input convex neural networks (ICNNs). We then use the networks to solve  optimal transport problems that require  a convex approximation of functions and demonstrate their effectiveness. Cubic-ICKANs produce results similar to those of ICNNs.
\end{abstract}

\section{Introduction}
Recently, Kolmogorov-Arnold networks (KANs) have been introduced as an alternative to multilayer perceptrons for high-dimensional function approximation, based on the Arnold-Kolmogorov representation theorem \cite{liu2024kan}. Arnold and Kolmogorov demonstrated \cite{kolmogorov1961representation} that a multivariate continuous smooth $\R$ valued function $f$ on a bounded domain can be expressed as a finite composition of sums of continuous single-variable functions. Specifically, if $f$ is continuous on $[0,1]^n$, then 
\begin{equation} 
f(x)= \sum_{i=1}^{2n+1} \psi_i\left(\sum_{j=1}^n \Phi_{i,j}(x_j)\right), \label{eq:kom} 
\end{equation} 
where $\Phi_{i,j}: [0,1] \longrightarrow \mathbb{R}$ and $\psi_i : \mathbb{R} \longrightarrow \mathbb{R}$, $i=1,\ldots,2n+1$, $j=1,\ldots,n$.
Since the one-dimensional functions $\psi$ and $\Phi$ can be highly irregular or even fractal, it has been demonstrated that they may not be practically learnable \cite{girosi1989representation,poggio2020theoretical}. To address this issue, K\r{u}rkov\'{a} in \cite[Theorem 1]{kurkova1992} proposed considering an approximate representation $f_m$ instead of an exact one for a continuous function $f$, also extending the outer sum to $m$ terms instead of $2n+1$, with $m$ intended to be large and greater than $2n+1$. 
More recently, under the assumption that $f$ is $\alpha$-Hölder ($0 < \alpha \leq 1$) and also using $m$ terms on the outer summation, \cite{song2025explicit} constructed an approximation $f_m$ with $\Phi_{i,j}$ $C^2$ converging to $f$ in the sup norm as $m$ increases (see also \cite{demb2021note} for another construction approximating the function in the sup norm).
In this work, we follow the approach of \cite{liu2024kan}. Firstly, they suggest not restricting the outer sum in \ref{eq:kom} to $2n+1$ terms but to $m$ terms and define a KAN $l^{th}$ layer, $l=$ $0,\ldots,L$, as an operator $\psi^l_{m,q}$ from $[0,1]^{m}$ to $\mathbb{R}^{q}$:
\begin{equation}
  \label{eq:PK1L}
  (\psi^l_{m,q}(x))_k=  \sum_{j=1}^m \Phi_{l,k,j}(x_j), \text{ for } k=1, \ldots, q.
\end{equation}
Second, by stacking the layers, i.e., composing the operators $\psi^l$, $l=0\ldots,L$, they define the KAN operator from $[0,1]^m$ to $\mathbb{R}$:
\begin{equation}
        K(x)= \psi^L_{n_{L-1},d} \circ \psi^{L-1}_{n_{L-2}, n_{L-1}} \circ \ldots \circ \psi^1_{n_0,n_1} \circ \psi^0_{m,n_0} (x).
    \label{eq:KAN}
\end{equation}
Since all functions $\Phi_{l,k,j}$ are one-dimensional, many classical numerical techniques are available to construct implementable approximations. The implementation proposed in \cite{liu2024kan} uses B-splines and is numerically costly, but other approximations based on wavelets \cite{bozorgasl2024wav}, radial basis functions \cite{li2024kolmogorovarnold,ta2024bsrbf}, and Chebyshev polynomials \cite{ss2024chebyshev} have been introduced to reduce computation time.

All these representations share the same limitation: the output of a layer may not lie on the grid originally chosen for the next layer. Mapping the output, a priori in $\mathbb{R}^q$, back to $[0,1]^q$ using, for example, a sigmoid function is possible, but this typically reduces the accuracy and slows convergence. Although \cite{liu2024kan} propose an adaptation technique, it fails numerically. Recently, using a P1 finite-element formulation, \cite{warin2024p1} introduced a new P1-KAN architecture that avoids this issue and established its convergence; numerical examples on function approximation using a mean squared error (MSE) criterion and real applications such as hydraulic valley optimization show that P1-KAN is more efficient than MLPs and all tested KAN variants.

\medskip

In this article, we extend the work of \cite{warin2024p1} and propose new architectures to approximate convex functions using convex approximations. Convexity is essential in many applications. For instance, in optimal transport, the optimal transport map is the gradient of a convex function (Brenier's theorem \cite{brenier1991}). Similarly, in optimal control, the value function may be convex or concave. For example, in gas storage optimization, it is crucial for practitioners to preserve the concavity of the Bellman value with respect to the stock level; the marginal cost, defined as the derivative of this Bellman value, must decrease as the stock level increases.

The approximation of a convex function while preserving convexity has been widely studied. Approximation by cuts using regression has been explored in \cite{balazs2015near,ghosh2019max,ghosh2021max,kim2024max}, leading to max-affine approximations. A max-affine representation using group-max neural networks \cite{warin2023groupmax} has been proposed and proved convergent, relying on the fact that a convex function can be $\epsilon$-approximated by the maximum of finitely many affine functions. Other theoretical results appear in \cite{calafiore2019log}, which uses a one-layer feedforward network with exponential activations inside and a log-sum-exp output, a standard convex function. Numerically, \cite{amos2017input} introduced input convex neural networks (ICNNs), whose convergence was proved in \cite{chen2018optimal}; ICNNs have since been used in many convex-approximation contexts, such as optimal transport \cite{makkuva2020optimal,korotin2019wasserstein}, optimal control \cite{chen2018optimal,agrawal2020learning}, inverse problems \cite{mukherjee2020learned}, and general optimization \cite{chen2020input}.  
Input-convex KANs have also been tested for modeling hyperelastic materials \cite{thakolkaran2025can}, independently of our work.

\medskip

We propose two Kolmogorov–Arnold architectures:
\begin{itemize}
\item The first version uses a piecewise linear approximation of the 1D functions in the KAN representation. It is based on the P1-KAN network of \cite{warin2024p1}. Unlike other KAN architectures, it takes as input a grid $G^1$ defining the approximation domain and outputs both the function value and a grid $G^2$ representing the image of $G^1$. This explicit definition facilitates layer composition. By enforcing convexity in the 1D approximation, we develop a new network and provide a Universal Approximation Theorem.

\item The second version uses a Hermite cubic spline approximation, also enforcing convexity. This network provides a higher-order approximation, which is desirable when the gradient of the convex function is of interest. In such cases, a piecewise linear approximation produces only piecewise-constant gradients, which may be insufficient. However, no convergence guarantees are established.
\end{itemize}

In \ref{sec:ICKAN}, we describe the two networks in the class of input convex  Kolmogorov–Arnold networks (ICKANs), giving both the adapted and non-adapted grid versions (as in \cite{warin2024p1}). We compare these variants under a mean squared error criterion and show that their performance is comparable to ICNNs.  
In \ref{sec:PICKAN}, we extend these networks to the case in which the function is convex only with respect to part of the input, yielding partial input convex  Kolmogorov–Arnold networks (PICKANs). We compare their performance with the corresponding partial ICNNs (PICNNs).  
Finally, in \ref{sec:optim_transport}, we use ICKANs to compute optimal transport maps between two distributions on simulated data, illustrating the efficiency of the proposed networks.

\section{ICKAN}
\label{sec:ICKAN}
We assume throughout this section that the target function is fully convex. 
In the first part, we explain how to construct a one-dimensional convex approximation of a function on $[0,1]$, either using a piecewise-linear representation or a cubic-spline one. In the second part, we detail the construction of the layers associated with the two proposed networks and we establish a Universal Approximation Theorem for the first architecture.  
In the third part, we present several numerical experiments.

\subsection{Approximation of a 1D convex function $f$ on $[0,1]$}
We introduce the lattice $ \{\hat{x}_{0}:=0\} \cup (\hat{x}_{p})_{1 \le p \le P-1} \cup  \{ \hat{x}_{P}:=1\}$ where the points $\hat{x}_p \in [0,1]$ form an increasing sequence indexed by $p$. 
\subsubsection{Piecewise linear approximation}
\label{sec:P1-1D}
\begin{figure}[tbhp]
    \centering
    \includegraphics[width=0.4\linewidth]{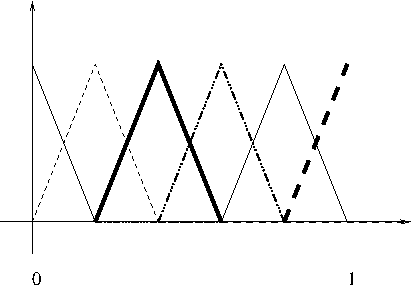}
    \caption{Uniform P1 basis functions on $[0,1]$ with $P=5$.}
    \label{fig:P1}
\end{figure}

For $P >0$, the degrees of freedom of a $P1$ approximation $ \phi$ of $f$ are the values of the function on the lattice leading to 
\begin{equation}
\phi(x) = \sum_{p=0}^{P} a_p \Psi_p(x)
   \label{eq:P1}
\end{equation}
where $(a_p)_{p=0,\ldots,P}$ are approximations of $(f( \hat{x}_p))_{p=0, \ldots, P}$ and $(\Psi_p)_{p=0,\ldots,P}$ is the basis of the shape functions: these functions have compact support in each interval $[\hat{x}_{p-1}, \hat{x}_{p+1}]$ for $p=1, \ldots, P-1$ and are defined as:
\begin{equation}
    \Psi_p(x) = \left \{
    \begin{array}{cc}
    \frac{x - \hat{x}_{p-1}}{\hat{x}_{p} - \hat{x}_{p-1}} &  \mbox{ for } x \in [ \hat{x}_{p-1} , \hat{x}_{p}], \\
    \frac{\hat{x}_{p+1} - x}{\hat{x}_{p+1} - \hat{x}_{p}} & \mbox{ for } x \in [ \hat{x}_{p} , \hat{x}_{p+1}], \\    
    \end{array}
    \right. 
    \label{eq:shape}
\end{equation}
for $p=1, \ldots, P-1$ and $\Psi_0(x) =  \max(1 - \frac{x-\hat{x}_0}{\hat{x}_1-\hat{x}_0},0)$, $\Psi_P(x) =  \max(\frac{x-\hat{x}_{P-1}}{\hat{x}_P-\hat{x}_{P-1}},0)$ (see \ref{fig:P1}).
Since we seek a convex function, the piecewise-linear approximation must have non-decreasing derivatives on each mesh interval $[\hat{x}_p,\hat{x}_{p+1}]$ for $p=0,\ldots,P-1$.
The trainable variables of the convex approximation are $b$, an approximation of $f(0)$,  $\hat b$, an approximation of $f'(0)$, and $( d_{i})_{i=1, \ldots, P-1}$ an approximation of $f'(\hat x_i)-f'(\hat x_{i-1})$ for $i=1, \ldots, P$.\\
Therefore $a_p \approx f(\hat x_p)$ in \ref{eq:P1} is given by:
\begin{equation}
    a_p=  b +   \sum_{j=1}^{p} \Bigl( \hat b +  \sum_{i=1}^{j-1} \max( d_i,0)\Bigr) \left(\hat x_{j} - \hat x_{j-1} \right),\,p=0, \ldots, P,
    \label{eq:approxDer}
\end{equation}
where the $\max$ is here to ensure that the approximation $f'( \hat x_p) \approx \hat b +  \sum_{i=1}^{p} \max( d_i,0)$ is non-decreasing in $p$.\\
It is also possible to adapt the grid, following \cite{warin2024p1}.  
In that case, the interior vertices in $]0,1[$ are initialized uniformly at random and trained to adapt to the data. This is achieved by defining, for $1\le p < P$,
\begin{align}
\label{eq:disPoint}
   \hat{x}_p = \frac{\sum_{k=1}^p e_k}{\sum_{k=1}^P e_k},
\end{align}
where $e_1, \ldots, e_P$ are positive  variables. 
Thus, $b$, $\hat{b}$, $(d_i)_{i=1,\ldots,P-1}$ and $(e_i)_{i=1,\ldots,P}$ are learned during training. In \ref{appendix:0_1}, we demonstrate the effect of adaptation on the P1-ICKAN.

 \subsubsection{The cubic approximation}
On a mesh  interval $[\hat{x}_{p}, \hat{x}_{p+1}]$, the function $f$ is approximated for $x \in [\hat{x}_{p},\hat{x}_{p+1}]$ using a cubic Hermite spline of the form  $\phi_p\left(\frac{x- \hat{x}_{p}}{\hat{x}_{p+1} -\hat{x}_{p}}\right)$ where:
\begin{align*}
    \phi_p(t)=& f(\hat{x}_{p})  h_{00}(t)+ f'(\hat{x}_{p}) h_{10}(t) (\hat{x}_{p+1} -\hat{x}_{p}) \\
    &+ f(\hat{x}_{p+1}) h_{01}(t) + f'(\hat{x}_{p+1}) h_{11}(t)  (\hat{x}_{p+1} -\hat{x}_{p})
\end{align*}
 and the cubic Hermite basis functions are given by
\begin{align*}
    h_{00}(t)= &2t^3-3t^2+1,\\
    h_{10}(t)= &t^3 -2 t^2 +t, \\
    h_{01}(t)= & -2t^3+3t^2, \\
    h_{11}(t)=&  t^3-t^2.
\end{align*}
As before, we require the derivatives to be non-decreasing. Under this condition, the values
$(f(\hat{x}_p))_{p=0,\ldots,P}$ must satisfy, for $p=0,\ldots,P-1$,
\begin{align*}
    &f(\hat{x}_{p}) + \frac{\hat{x}_{p+1} -\hat{x}_{p}}{3} \Bigl(2f'(\hat{x}_{p}) +f'(\hat{x}_{p+1})\Bigr) \le f(\hat{x}_{p+1}),\\
     &f(\hat{x}_{p+1}) \le f(\hat{x}_{p}) + \frac{\hat{x}_{p+1} -\hat{x}_{p}}{3} \Bigl(f'(\hat{x}_{p}) + 2f'(\hat{x}_{p+1})\Bigr). 
\end{align*}
Therefore the convex  cubic-spline approximation on each mesh interval $[\hat{x}_{p},\hat{x}_{p+1}]$ for $p=0,\ldots, P-1$ is given by $\phi_p\left(\frac{x- \hat{x}_{p}}{\hat{x}_{p+1} -\hat{x}_{p}}\right)$ with
\begin{equation}     \label{eq:cubic1D}
    \phi_p(t) = a^{0}_p  h_{00}(t)+ a^{1}_p h_{10}(t) (\hat{x}_{p+1} -\hat{x}_{p}) + a^{0}_{p+1} h_{01}(t) + a^{1}_{p+1}  h_{11}(t)  (\hat{x}_{p+1} -\hat{x}_{p}),
\end{equation}
 where :
\begin{equation}\label{eq:cubic1DExp}
\begin{split}
a^1_p= &  \hat b +  \sum_{i=1}^{p} \max( d_i,0), \\
a^0_p= &    b + \sum_{i=1}^{p} \frac{\hat{x}_{i} -\hat{x}_{i-1}}{3} \Bigl(2 a^1_{i-1} + a^1_{i}  +\sigma(g_i) (a^1_{i} -a^1_{i-1})\Bigr),
\end{split}
\end{equation}
$\sigma$ denotes the sigmoid function, and the parameters  $\hat b, b, (d_i)_{i=1,\ldots, P}, (g_i)_{i=1,\ldots, P}$ are trainable.
In the adapted version, the variables $(e_i)_{i=1,\ldots,P}$ determining the grid positions 
$(\hat{x}_p)_{p=1,\ldots,P-1}$ are also trained, exactly as in the piecewise-linear case. In \ref{appendix:0_2}, we demonstrate the effects of adapting Cubic-ICKAN and optimising one layer.

\subsection{The ICKAN layers}
As in the P1–KAN architecture, an ICKAN layer takes as input a hypercube defining 
the domain together with the batched values $x$, and outputs both an estimate of 
the function at $x$ and a hypercube representing the image of the original domain. 
This structure ensures that the layers can be composed in a consistent manner.

\subsubsection{The P1-ICKAN layers}
For the first component of the first layer, we define $\hat \kappa^0_{n,m}$ for $x \in   \left[0,1\right]^n $ and the hypercube $G^0 := [0,1]^n$ as:
\begin{equation}  \label{eq:firstLPart}
\begin{split}
  \hat \kappa^0_{n,m}&(x, G^0)_k =\\
  &\sum_{j=1}^{n} \sum_{p=0}^{P}   \biggl(b_{0 ,k,j} +   \sum_{s=1}^{p} \Bigl( \hat b_{0,k,j} +  \sum_{i=1}^{s-1} \max( d_{0,k,j,i},0)\Bigr) (\hat x_{0,j,s} - \hat x_{0,j,s-1})\biggr)\Psi^{0,j}_{p}(x_{j}),\\
  &\text{ for }  k=1, \ldots, m.
\end{split}
\end{equation}

The values $(\hat{x}_{0,j,s})_{s=0,\ldots,P}$ denote the one-dimensional grid 
for the $j$\textsuperscript{th} dimension of $G^0$, and $\Psi^{0,j}_p$ for $p=1,\ldots,P$ are the shape functions 
defined in \ref{eq:shape} and shown in \ref{fig:P1}.\\
The image $G^1 = \displaystyle{\prod_{k=1}^m} [\underline{G}^1_k,\bar{G}^1_k ] $ of $G^0$ by $\hat \kappa^0_{n,m}$ is exactly given by:
\begin{equation}
\label{eq:grid}
\begin{split}
    \underline{G}^1_k = &\sum_{j=1}^{n} \min_{0 \le p \le P} \left[b_{0 ,k,j} +   \sum_{s=1}^{p} \Bigl( \hat b_{0,k,j} +  \sum_{i=1}^{s-1} \max( d_{0,k,j,i},0)\Bigr) (\hat x_{0,j,s} - \hat x_{0,j,s-1}) \right] \\
    \bar{G}^1_k = & \sum_{j=1}^{n}  \max \left[ b_{0 ,k,j} , b_{0 ,k,j} +   \sum_{s=1}^{P} \Bigl( \hat b_{0,k,j} +  \sum_{i=1}^{s-1} \max( d_{0,k,j,i},0)\Bigr) (\hat x_{0,j,s} - \hat x_{0,j,s-1})  \right]
    \end{split}
\end{equation}
for $1 \le k \le m$.
Thus, the layer output is defined as
\begin{equation}
    \kappa^0_{n,m}(x, G^0) = (\hat \kappa^0_{n,m}(x, G^0), G^1).
    \label{eq:firstLayer}
\end{equation}
\begin{remark}
Since the approximation is convex, the maximum $\bar{G}^1_k$ is attained on 
the boundary of the domain.
\end{remark}
As a sum of convex one-dimensional functions in each coordinate, 
$\hat{\kappa}^0_{n,m}(\cdot,G^0)$ has a diagonal, positive-definite Hessian.

\medskip

To ensure convexity is preserved through composition, 
we require each subsequent layer $\hat{\kappa}^l_{m,q}(x,G^l)$ ($l\ge1$) 
to be convex and non-decreasing in $x$ as in \cite{amos2017input}. It suffices to impose 
$\hat{b}_{l,k,j} \ge 0$ for all $k$ and $j$.
Thus, for $l\ge1$, with $G^l$ the input hypercube, we define
\begin{equation*}
    \begin{split}
  &\hat \kappa^l_{m,q}(x, G^l)_k =\\ 
  &\sum_{j=1}^{m} \sum_{p=0}^{P}   \biggl(b_{l ,k,j} +   \sum_{s=1}^{p} \Bigl(  \max(\hat b_{l,k,j},0)+  \sum_{i=1}^{s-1} \max( d_{l,k,j,i},0)\Bigr) (\hat x_{l,j,s} - \hat x_{l,j,s-1}) \biggr) \Psi^{l,j}_{p}(x_{j}), \\
 &  \text{ for }  k=1, \ldots, q,
  \end{split}
\end{equation*}
and the output hypercube $G^{l+1}$ image of $G^l$ is defined analogously to \ref{eq:grid}.
The complete layer is then defined as
\begin{equation}
    \kappa^l_{m,q}(x, G^l) = (\hat \kappa^l_{m,q}(x, G^l), G^{l+1}).
    \label{eq:secondLayer}
\end{equation}
By concatenating $L$ layers, and denoting $n_l$ the number of neurons in layer $l$
(as in \ref{eq:KAN}), assuming the number of mesh points $P$ is fixed across neurons,
we define the full network:
\begin{equation}
    K(x)= \hat \kappa^{L}_{n_{L-1},1} \circ  \kappa^{L-1}_{n_{L-2},n_{L-1}}  \circ \ldots  \circ  \kappa^{0}_{n,n_0}(x, [0,1]^n).
    \label{eq:KConc}
\end{equation}
As in \cite{warin2024p1}, we propose two variants:
\begin{itemize}
\item \textbf{Uniform grid.}  
The mesh is fixed, and the trainable parameters are
    \begin{equation*}
 \begin{split}
     \mathcal{A} :=&  (b_{0,k,j}, \hat b_{0,k,j}, (d_{0,k,j,i})_{i=1, \ldots, P-1})_{k=1,\ldots ,n_0, j=0, \ldots, n} \cup \\
     & (b_{l,k,j}, \hat b_{l,k,j}, (d_{l,k,j,i})_{i=1, \ldots, P-1})_{l=1, \ldots, L-1, k=1,\ldots ,n_l, j=0, \ldots ,n_{l-1}} \cup \\
     &(b_{L,1,j}, \hat b_{L,1,j}, (d_{L,1,j,i})_{i=1, \ldots, P-1})_{ j=0, \ldots ,n_{L-1}}.\\
    \end{split}
    \end{equation*}
    \item \textbf{Adaptive grid.}  In addition to $\mathcal{A}$, the interior grid points are trainable, i.e.
    \begin{equation*}\mathcal{A} \cup  (e_{0,j,p})_{j=1,\ldots,n, p=1, \ldots, P} \cup  ( e_{l,j,p})_{l=1,\ldots, L, j=1,\ldots, n_l, p=1, \ldots, P}  
    \end{equation*}
    where the variables $e_{l,j,p}$ determine the grid values 
$\hat{x}_{l,j,p}$ via a formula similar to \ref{eq:disPoint}.
\end{itemize}
\subsubsection{The Cubic-ICKAN layers} 
Consider again the one-dimensional lattice $(\hat{x}_{0,j,s})_{s=0,\ldots,P}$ on 
$G^0 := [0,1]^n$ associated with the $j$th input dimension. 
We define $\hat{\kappa}^0_{n,m}$ for
\[
x \in \prod_{j=1}^{n} [\hat{x}_{0,j,p_j}, \hat{x}_{0,j,p_j+1}],
\qquad p_j \in \{0,\ldots,P-1\},\; j=1,\ldots,n,
\]
and for the hypercube $G^0$ by
\begin{equation}
\begin{split}
  \hat \kappa^0_{n,m}(x, G^0)_k =&  \sum_{j=1}^{n}    a^0_{0,j,p_j}  h_{00}(t_{0,j})+ a^{1}_{0,j,p_j} h_{10}(t_{0,j}) (\hat{x}_{0,j,p_j+1} -\hat{x}_{0,j,p_j})   \\
  &+ a^{0}_{0,j,p_j+1} h_{01}(t_{0,j}) + a^{1}_{0,j,p_j+1}  h_{11}(t_{0,j})  (\hat{x}_{0,j,p_j+1} -\hat{x}_{0,j,p_j}),\\
  &\text{ for }  k=1, \ldots, m,
  \end{split}
  \label{eq:firstLPartCubic}
\end{equation}
where $t_{0,j} = \frac{x_j - \hat x_{0,j,p_j}}{\hat x_{0,j,p_j+1}-\hat x_{0,j,p_j}}$,
\begin{equation}\label{eq:Cubic-param1}
\begin{split}
&a^1_{0,j,p_j}=  \hat b_{0,j} +  \sum_{i=1}^{p_j} \max( d_{0,j,i},0),\\
&a^0_{0,j,p_j}=   b_{0,j} + \sum_{i=1}^{p_j} \frac{\hat{x}_{0,j,i} -\hat{x}_{0,j,i-1}}{3} \left(2 a^1_{0,j,i} + a^1_{0,j,i-1}  +\sigma(g_{0,j,i}) (a^1_{0,j,i} -a^1_{0,j,i-1})\right).
\end{split}
\end{equation}
The image hypercube 
 $G^1 = \displaystyle{\prod_{k=1}^m} [\underline{G}^1_k,\bar{G}^1_k ]$  is defined by
 \begin{equation}
\label{eq:gridCubic}
\begin{split}
    \underline{G}^1_k = &\sum_{j=1}^{n} \min_{0 \le p \le P} a^0_{0,j,p} \\
    \bar{G}^1_k = & \sum_{j=1}^{d_0}  \max \left[ a^0_{0,j,0}, a^0_{0,j,P} \right].\\
    \end{split}
\end{equation}
Note that $G^1$ is itself included in the image of $G^0$ under $\hat{\kappa}^0_{n,m}$.

To ensure that $G^1$ coincides exactly with the image of the layer, we may truncate the output:
\begin{equation*}
    \kappa^0_{n,m}(x, G^0) = \Bigl( \Bigl(\max\bigl(\hat \kappa^0_{n,m}(x, G^0)_k,\underline{G}^1_k\bigr)\Bigr)_{k=1,\ldots,m}, G^1\Bigr).
\end{equation*}
\begin{remark}
Due to convexity, only the minimum value needs to be truncated.
Clipping the value function reveals the effective image of the layer, 
which can then be used as the domain for the next layer. 
Semi-Lagrangian schemes in control problems also perform clipping inside interpolations, 
yielding nearly monotone schemes with provable convergence rates \cite{warin2016some}.  
In practice, the original KAN architecture does not adjust its grid to the output domain 
and still performs well; numerically, the effect of the truncation here is small. 
\end{remark}
Similarly to the piecewise linear case, for $l\ge 1$, $G^l$ being the hypercube defining the domain, the operator $\hat \kappa^l_{m,q}(x, G^l)_k$ for $l\geq 1$ is defined similarly to \ref{eq:firstLPartCubic} using
\begin{equation*}
a^1_{l,j,p_j}=   \max(\hat b_{l,j},0) +  \sum_{i=1}^{p_j} \max( d_{l,j,i},0) 
\end{equation*}
instead of  \ref{eq:Cubic-param1}. The output hypercube $G^{l+1}$ image of $G^l$ is given by an expression similar to \ref{eq:gridCubic}. Then the layer is defined by:
\begin{equation*}
    \kappa^l_{m,q}(x, G^l) = \Bigl(\Bigl(\max\bigl(\hat \kappa^l_{m,q}(x, G^l)_k,\underline{G}^{l+1}_k\bigr)\Bigr)_{k=1,\ldots,q} , G^{l+1}\Bigr).
\end{equation*}
The concatenation of layers is still defined by \ref{eq:KConc}.  
As in the piecewise-linear case, one may use either a uniform grid or an adaptive grid.

\subsubsection{Convergence results}
We provide two universal approximation theorems in the piecewise-linear setting, 
corresponding to the adapted and non-adapted cases.
\begin{theorem}
The space spanned by the P1-ICKAN allowing $n_l$ for \\ $l=0, \ldots,L-1$ and $L$  to vary for $P>1$ is dense in the set of Lipschitz convex functions on $[0,1]^n$ with respect to the sup norm when adaptation is used.
\label{theo:universelAdapt}
\end{theorem}
The proof of \ref{theo:universelAdapt} is given in \ref{appendix:proof1}.
We also establish a theorem for the non-adapted case; the proof is provided in 
\ref{appendix:proof2}.
 \begin{theorem}
    The space spanned by the P1-ICKAN  allowing  $n_l$ for \\ $l=0, \ldots,L-1$, $L$ and $P$ to vary is dense in set of Lipschitz convex functions on $[0,1]^n$ with respect to the sup norm when no adaptation is used.\label{theo:universelNoAdapt}
\end{theorem}

\subsubsection{Numerical Results} 
\label{sec:ickan_num} 
In this section, we compare the performance of the proposed networks with that of the ICNN. 
We first study their approximation capabilities on a $d$-dimensional convex function, 
and then examine a two-dimensional toy control example.

\medskip
\paragraph{\bf Numerical approximation}
We estimate the function
 \begin{equation}
    \label{eq:f_cubic}
     f(x)= \dps{\sum_{i=1}^d} (|x_i| + |1-x_i|) + x^{\top} A x
 \end{equation}
 where $A$ is a positive-definite matrix, using both an input convex  neural network 
with ReLU activation and an input convex  Kolmogorov network approximation 
$\tilde f^{\theta}$ parametrized by~$\theta$. The approximation is obtained by minimizing the empirical version of
\begin{equation}
\E[ (f(X) -\tilde f^\theta(X))^2]
    \label{eq:mse}
\end{equation}
where $X \sim U[-2,2]^d$.
We use the ADAM optimizer with a learning rate of $10^{-3}$, a batch size of $1{,}000$,
and perform $200{,}000$ iterations.  
Based on $10$ independent runs, we compute the average MSE and its standard deviation 
on a validation batch of size $100{,}000$ for different parameter choices.

For the ICNN, we consider networks with $2$, $3$, $4$, or $5$ layers, 
with a number of neurons chosen from $\{10, 20, 40, 80, 160, 320\}$, 
and ReLU activation. Among all tested configurations, the ICNN yielding the smallest 
average MSE is selected as the reference model.

Results for dimensions $3$ and $7$ are reported in 
\ref{tab:compICNN-ICKAN3,tab:compICNN-ICKAN7}. For the P1–ICKAN, the performance is stable across values of 
$P$ and across the number of layers.  
The adaptive version (P1–ICKAN adapt) performs better than the non-adaptive one.  
In dimension $3$, the adaptive P1–ICKAN achieves accuracy comparable to the 
best ICNN, while in dimension $7$ it outperforms the ICNN, despite having 
significantly fewer parameters.

Interestingly, the computational cost is almost independent of $P$ and of the 
problem dimension, thanks to efficient GPU parallelization. Although the P1–ICKAN uses fewer parameters than a feedforward 
network of comparable accuracy, the computational time is typically three to 
four times larger.

\begin{table}[tbhp]
    \centering
        \caption{Results over 10 runs  for the  minimization of \ref{eq:mse} in dimension 3 for the  function in \ref{eq:f_cubic} : ICCN versus P1-ICKAN. {\it NBL} stands for the number of layers, {\it NBN} the number of neurons, {\it nbParam} the number of model parameters, while {\it Time} stands for the time in seconds for one hundred iterations of the stochastic gradient descent.} 
    \label{tab:compICNN-ICKAN3}
    \begin{tabular}{|c|c|c|c|c|c|c|c|} \hline
    method & NBL & NBN & $P$ & Average & std  & nbParam & Time\\ \hline
    Best ICNN  & 2 & 320 &  & 9.37E-05 & 8.29E-05 &  105284& 0.14 \\ \hline
P1-ICKAN & 2 & 20 & 20 & 1.56E-04 & 7.39E-05  & 10080  & 0.54 \\ 
P1-ICKAN & 2 & 20 & 40 & 1.14E-04 & 9.79E-05  & 19680   &  0.54\\ 
P1-ICKAN   & 2 & 40 & 20 & 2.62E-04 & 3.68E-04  & 36960  & 0.55\\ 
P1-ICKAN   & 2 & 40 & 40 & 2.69E-04 & 4.65E-04   &  72160 & 0.53 \\ 
P1-ICKAN   & 3 & 20 & 20 & 3.13E-04 & 3.10E-04  & 18480  &  0.73\\ 
P1-ICKAN   & 3 & 20 & 40 & 6.83E-04 & 1.04E-03  &  36080 & 0.73\\ 
P1-ICKAN   & 3 & 40 & 20 & 2.76E-04 & 1.90E-04  & 70560  & 0.73\\ 
P1-ICKAN   & 3 & 40 & 40 & 1.67E-04 & 2.91E-04  & 137760 & 0.73 \\ \hline
P1-ICKAN adapt  & 2 & 20 & 20 & 1.23E-04 & 4.62E-05  & 10940 & 0.66\\ 
P1-ICKAN adapt  & 2 & 20 & 40 & 1.93E-04 & 3.72E-04  &21400 & 0.66\\ 
P1-ICKAN adapt  & 2 & 40 & 20 & 2.62E-04 & 2.97E-04  &38620 & 0.66 \\ 
P1-ICKAN adapt  & 2 & 40 & 40 & 1.13E-04 & 1.42E-04  & 75480& 0.66\\ 
P1-ICKAN adapt  & 3 & 20 & 20 & 9.82E-04 & 2.23E-03  & 19740 & 0.88\\ 
P1-ICKAN adapt  & 3 & 20 & 40 & 1.93E-04 & 2.75E-04  &38600 & 0.88 \\ 
P1-ICKAN adapt  & 3 & 40 & 20 & 3.63E-04 & 6.50E-04  & 73020& 0.88\\ 
P1-ICKAN adapt  & 3 & 40 & 40 & 5.35E-04 & 1.16E-03  & 142680& 0.88\\ \hline
    \end{tabular}
\end{table}

\begin{table}[tbhp]
    \centering
    \caption{Results over 10 runs for the minimization of \ref{eq:mse} in dimension 7 for function in \ref{eq:f_cubic}: ICCN versus P1-ICKAN. {\it NBL} stands for the number of layers, {\it NBN} the number of neurons, {\it nbParam} the number of model parameters, while {\it Time} stands for the time in seconds for one hundred iterations of the stochastic gradient descent.}
    \label{tab:compICNN-ICKAN7}
    \begin{tabular}{|c|c|c|c|c|c|c|c|} \hline
    method & NBL &  NBN & $P$  & Average & std   & nbParam & Time\\ \hline
Best ICNN  & 2 & 320 &  & 2.39E-03 & 6.72E-04 & 107848 & 0.14 \\ \hline
P1-ICKAN   & 2 & 40 & 10 & 8.19E-03 & 9.23E-04 & 21120 & 0.54\\ 
P1-ICKAN   & 2 & 40 & 20 & 3.71E-03 & 4.69E-03 & 40320&  0.54\\ 
P1-ICKAN   & 2 & 40 & 40 & 1.94E-03 & 1.20E-03 &  78720& 0.53\\ 
P1-ICKAN   & 3 & 40 & 10 & 9.05E-03 & 1.25E-03  &38720 & 0.73\\ 
P1-ICKAN   & 3 & 40 & 20 & 2.40E-03 & 8.88E-04  & 73920& 0.73\\ 
P1-ICKAN   & 3 & 40 & 40 & 3.45E-03 & 2.09E-03  & 144320& 0.73\\ \hline
P1-ICKAN adapt  & 2 & 40 & 10 & 2.31E-03 & 6.88E-04 & 21990 & 0.66\\ 
P1-ICKAN adapt  & 2 & 40 & 20 & 2.72E-03 & 2.95E-03 & 42060& 0.67\\ 
P1-ICKAN adapt  & 2 & 40 & 40 & 1.33E-03 & 1.03E-03  &82200 & 0.67\\ 
P1-ICKAN adapt  & 3 & 40 & 10 & 2.69E-03 & 1.08E-03 & 39990& 0.88\\ 
P1-ICKAN adapt  & 3 & 40 & 20 & 2.87E-03 & 1.89E-03 & 76460& 0.89\\ 
P1-ICKAN adapt  & 3 & 40 & 40 & 2.24E-03 & 2.47E-03  &149400 & 0.89 \\ \hline
    \end{tabular}

\end{table}

    In \ref{tab:compICNN-CUBICICKAN7}, results in dimension $7$ using the  Cubic-ICKAN show that the approximation is good but at a computational cost much higher than the P1-ICKAN.
\begin{table}[tbhp]
    \centering
    \caption{Results over 10 runs  for the  minimization of \ref{eq:mse} in dimension 7 for function in \ref{eq:f_cubic}: ICCN versus  Cubic-ICKAN (C-ICKAN). {\it NBL} stands for the number of layers, {\it NBN} the number of neurons, {\it nbParam}  the number of model parameters, while {\it Time} stands for the time in seconds for one hundred iterations of the stochastic gradient descent.} 
    \label{tab:compICNN-CUBICICKAN7}
    \begin{tabular}{|c|c|c|c|c|c|c|c|} \hline
    method & NBL &  NBN & $P$  & Average & std   & nbParam & Time\\ \hline

Best ICNN  & 2 & 320 &  & 2.39E-03 & 6.72E-04 & 107848 & 0.14 \\ \hline
C-ICKAN   & 2 & 20 & 10 & 6.66E-03 & 2.06E-03 &  12320& 1.32 \\ 
C-ICKAN   & 2 & 20 & 20 & 3.59E-03 & 7.64E-04 &  23520& 1.31 \\ 
C-ICKAN   & 2 & 40 & 10 & 3.88E-03 & 1.06E-03  & 42240  & 1.47\\ 
C-ICKAN   & 3 & 20 & 10 & 6.19E-03 & 2.11E-03 & 21120 & 1.79 \\ 
C-ICKAN   & 3 & 20 & 20 & 5.04E-03 & 2.81E-03 & 40320 & 1.79\\ 
C-ICKAN   & 3 & 40 & 10 & 3.82E-03 & 1.64E-03  & 77440 & 2.03\\ \hline
C-ICKAN adapt  & 2 & 20 & 10 & 1.43E-03 & 6.29E-04 &12790 &  1.49 \\ 
C-ICKAN adapt  & 2 & 20 & 20 & 1.19E-03 & 8.76E-04 &24460 & 1.55\\ 
C-ICKAN adapt  & 2 & 40 & 10 & 1.38E-03 & 1.28E-03  & 43110& 1.65\\ 
C-ICKAN adapt  & 3 & 20 & 10 & 3.06E-03 & 1.66E-03 & 21790& 2.06 \\ 
C-ICKAN adapt  & 3 & 20 & 20 & 1.95E-03 & 1.43E-03  & 41660& 2.07\\ 
C-ICKAN adapt  & 3 & 40 & 10 & 1.49E-03 & 1.11E-03  & 78710& 2.23\\  \hline
    \end{tabular}
\end{table}

\begin{figure}[tbhp]
     \centering
    \includegraphics[width=0.5\linewidth]{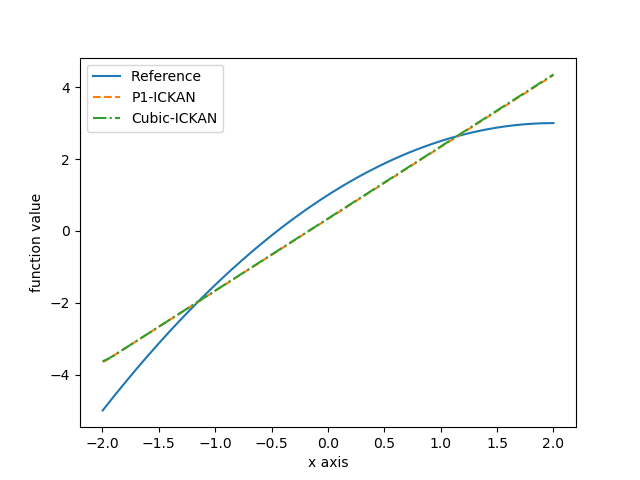}
     \caption{ Estimation of the function in~\ref{eq:wrong_Con} \rd{for dimension $d=1$} wrongly supposing that the function is convex, network with $P=10$.}
     \label{fig:nonConErr1D}
 \end{figure}

\begin{figure}[tbhp]
     \centering
     \begin{minipage}{0.49\linewidth}
    \includegraphics[width=\linewidth]{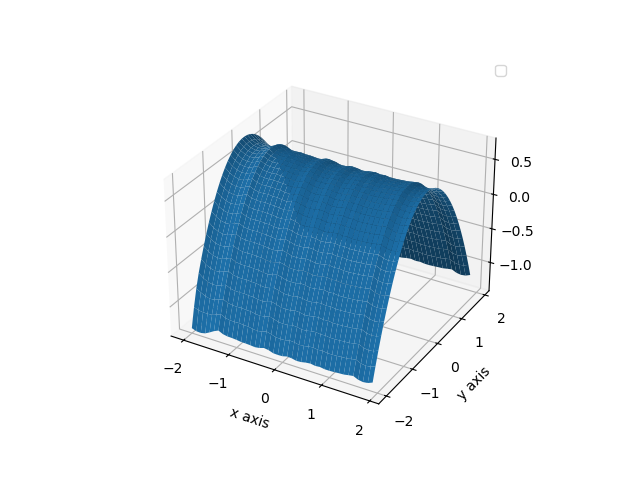}
     \caption*{P1-ICKAN }
     \end{minipage}
     \begin{minipage}{0.49\linewidth}
    \includegraphics[width=\linewidth]{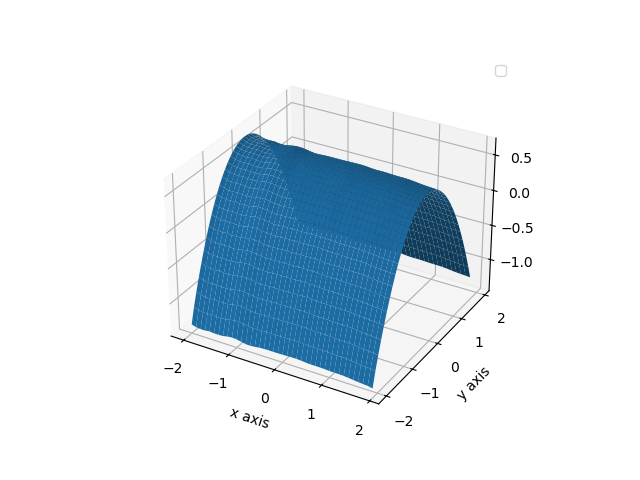}
     \caption*{Cubic-ICKAN }
     \end{minipage}
     \caption{Error for the function in~\ref{eq:wrong_Con} \rd{for dimension $d=2$} wrongly supposing that the function is convex, network with $P=10$.}
     \label{fig:nonConErr2D}
 \end{figure}

We may wonder what happens if we incorrectly assume that the function to be learned is convex.  
 We consider
\begin{equation}
\label{eq:wrong_Con}
     f(x)= A  + 2 {\bf 1}_d^\top x + x^\top Q x,
 \end{equation}
 where $Q=-0.5$ if $d=1$, $Q = \left ( \begin{array}{cc}
     1 & 0 \\
     0 & -0.5
 \end{array}\right)$ if $d=2$, ${\bf 1}_d$ the vector of dimension $d$ with all components equal to 1 \rd{and $A=1$}. \ref{fig:nonConErr1D}  \rd{shows the reference function and its approximation for the case $d=1$ while} \ref{fig:nonConErr2D} \rd{ shows the error as the difference between the real function and its approximation for the case $d=2$: it} shows that the error is concentrated on the axis supporting the \rd{non-}convexity. The quadratic shape of the error indicates that the network is clipping the negative eigenvalues of the hessian matrix.

\medskip

\paragraph{\bf A toy control example in dimension 2}
We consider the simple linear quadratic problem with dynamics
\begin{align*}
    x_{t+1} = A x_t + B u_t + w_t,\,t=0, \ldots, N-1,
\end{align*}
where $(x_t, u_t, w_t) \in \mathbb{R}^2 \times \mathbb{R}^2 \times \mathbb{R}^2$,  
$(w_t)_{t=0,\ldots,N-1}$ are i.i.d. centered Gaussian noises with covariance 
matrix $W \in \mathbb{R}^{2 \times 2}$ \rd{ and $x_0$ is given }.  
The control is chosen in feedback form $u_t = \phi_t(x_t)$ before observing the noise.  
The matrices $A,B \in \mathbb{R}^{2 \times 2}$ are given.  
The objective is to minimize
\begin{align*}
    J(x_0,\rd{(u_t)_{t=0, \ldots,N-1}})= \E\Bigl[ \sum_{t=0}^{N-1}  \rd{\Big (} x_t^\top Q x_t + u_t^\top R u_t \rd{\Big )}  +  x_N^\top  Q_f x_N | x_0\Bigr]
\end{align*}
where $Q$ and $R$ are two positive \rd{semi-definite} two-dimensional matrices.\\
The optimal control \rd{$u_t^*$ associated the optimal trajectory $x^*_t$} is given by
\begin{align*}
    \rd{u_t^*}=\phi_t(x_t)= & K_t x_t\rd{^*}, \\
    K_t=& - (B^\top P_{t+1} B +R)^{-1} B^\top P_{t+1} A,
\end{align*}
where the backward Riccati recursion is
\begin{align*}
    P_{t} = & A^\top P_{t+1} A - A^\top P_{t+1} B (B^\top P_{t+1} B +R)^{-1} B^\top P_{t+1} A +Q\\
    P_N= & Q_f
\end{align*}
and the \rd{optimal cost} function \rd{$J^*$} satisfies $\rd{J^*(x_0)}= x_0^\top P_0 x_0 +r_0$
with
\begin{align*}
    r_t= & r_{t+1} + \Tr{[W P_{t+1}]} \\
    r_N = & 0.
\end{align*}
We assume that the optimal control $u_t^*$ is known exactly, and our goal is to 
approximate the convex value function $J\rd{^*}$ on $[-3,3]^2$.  
Applying the optimal control to trajectories generated from a sample of 
$(w_t)_{t=0,\ldots,N-1}$ and $x_0 \sim U([-3,3]^2)$ yields the cost
\begin{align*}
    C(x_0,(w_t)_{t=0, \ldots,N-1}) = \sum_{t=0}^{N-1} \rd{\Big (} (x_t)^\top Q x_t + (K_t x_t)^\top R (K_t x_t)  \rd{\Big )}  +  x_N^\top  Q_f x_N.
\end{align*}
Using a convex neural network $\kappa^{\theta}$ parametrized by $\theta$ to approximate $J\rd{^*}$, 
we minimize
\begin{equation} \label{eq:opt_control}
    L(\theta) = \E[ (\kappa^\theta(x_0) -C(x_0,(w_t)_{t=0, \ldots,N-1}))^2 ]
\end{equation}
We choose $Q = R = A = B = Q_f = W = 
\begin{pmatrix}
1 & \rd{0} \\ \rd{0} & 1
\end{pmatrix}$ \rd{for our test case}.
The optimizer settings are the same as in the previous experiment.  
For the ICNN, we use three hidden layers with $30$ neurons.  
For the ICKAN, we use two hidden layers with $10$ neurons and a mesh size $P=10$.
One hundred iterations of stochastic gradient descent require $0.29$ seconds for the ICNN, 
$0.69$ seconds for the P1-ICKAN, and $1.14$ seconds for the Cubic-ICKAN.

\begin{figure}[tbhp]
    \centering
   \begin{minipage}{0.49\linewidth}
    \includegraphics[width=\linewidth]{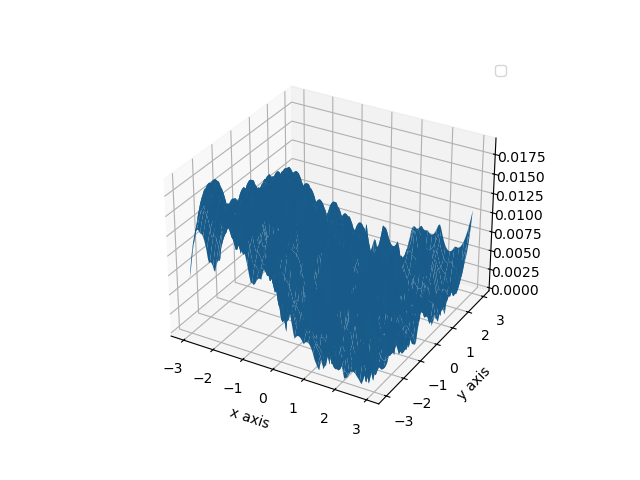}
     \caption*{ICNN }
     \end{minipage}
         \begin{minipage}{0.49\linewidth}
    \includegraphics[width=\linewidth]{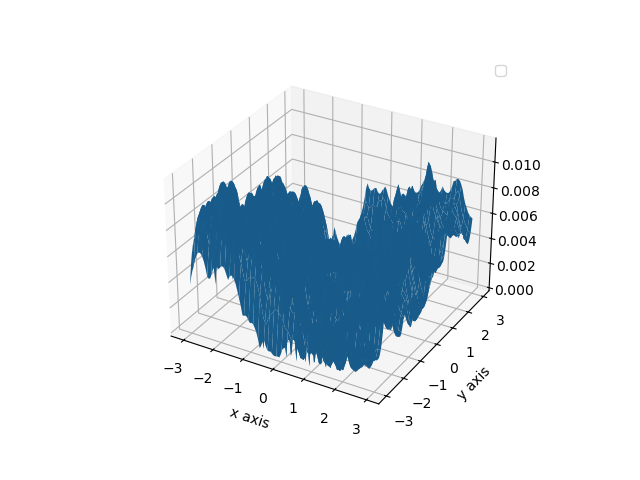}
     \caption*{P1-ICKAN }
     \end{minipage}
     \begin{minipage}{0.49\linewidth}
    \includegraphics[width=\linewidth]{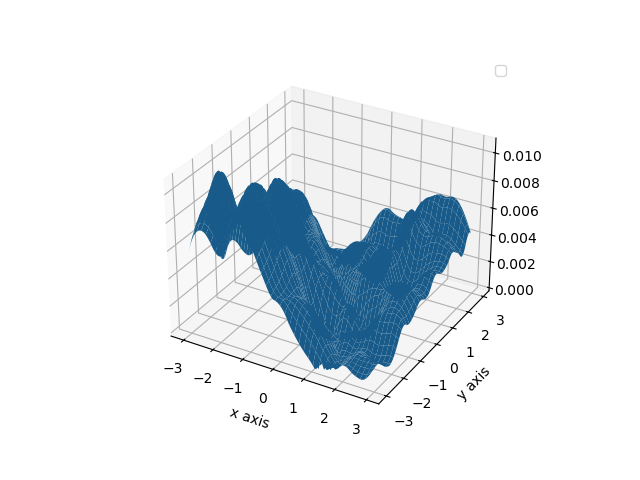}
     \caption*{Cubic-ICKAN }
     \end{minipage}  
    \caption{Relative error given by~\ref{eq:opt_control} obtained on the linear quadratic optimal control problem.}
    \label{fig:LQ}
\end{figure}

\ref{fig:LQ} displays the relative errors obtained by the three networks.  
The maximal error is reduced by nearly a factor of two when using convex KAN architectures compared with ICNN.
As expected, the cubic version yields a smoother approximation.

\section{PICKAN}
\label{sec:PICKAN}
We assume in this section that the function $f$ to be approximated is partially convex: 
$f(x,y)$ is convex in $y \in \mathbb{R}^{n_y}$ but not necessarily convex in 
$x \in \mathbb{R}^{n_x}$.  
We assume that the function is defined on 
$G^0 = G_x^0 \times G_y^0$,  
where $G_x^0 = [0,1]^{n_x}$ and $G_y^0 = [0,1]^{n_y}$.

We present the network architecture using a piecewise-linear approximation of the 
one‑dimensional functions, although the same construction can be adapted to a 
cubic-spline approximation.

We denote by P1–KANL the layer structure introduced in \cite{warin2024p1}, and by 
$\rho^l_{p,q}$ the corresponding operator for layer $l$ with input dimension $p$ 
and output dimension $q$.
The first layer $\kappa^0$, defined in \ref{eq:firstLayer}, is denoted ICKANL0, 
while any layer $\kappa^l$ with $l>0$, defined in \ref{eq:secondLayer}, is denoted ICKANL1.

\begin{figure}[tbhp]
    \centering
\includegraphics[width=0.9\linewidth]{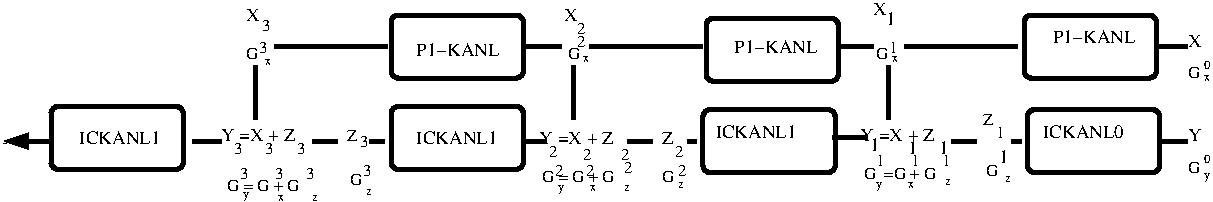}
    \caption{Partial input convex  Kolmogorov Arnold network using a piecewise linear approximation.}
    \label{fig:PICKAN}
\end{figure}

The architecture of the network is illustrated in \ref{fig:PICKAN}.
The recursion described in \ref{fig:PICKAN} is given for $M >0$  by:
\begin{align*}    
    (X_1, G^1_x) =& \rho^0_{n_x,M}(x, G^0_x),\\  
    (Y_1, G^1_y) = & (X_1, G^1_x) + \kappa^0_{n_y,M}(y,G^0_y), \\
    (X_{i+1},G^{i+1}_x) =& \rho^{i}_{M,M}(X_i, G^i_x), \\
    (Y_{i+1}, G^{i+1}_y)= & (X_{i+1}, G^{i+1}_x) + \kappa^{i}_{M,M}(Y_i, G^i_y)  \quad \mbox{  for } i =1, L-1, \\
(Y_{L+1}, G^{L+1}_y)= & \kappa^{L}_{M,1}(Y_{L}, G^{L}_y).
\end{align*}
\begin{remark}
When $f$ is non-convex, setting $\kappa^0_{n_y,M}=0$, $\kappa^l_{M,M} = 0$ for $1 <l < L$ and 
$\kappa^L_{M,1}(x) = x_1$ recovers the P1–KAN architecture, which is known to converge 
when the Kolmogorov–Arnold basis functions are Lipschitz.  
When the function $f$ is convex, forcing the P1–KANL outputs to vanish 
recovers the ICKAN network.
\end{remark}
\begin{remark}
In the PICKAN architecture, the contributions from the convex and non-convex parts 
are added together, in the same manner as in the Kolmogorov–Arnold layer 
(\ref{eq:PK1L}).
\end{remark}
\begin{remark}
Depending on the regularity of the function in $x$ and in $y$, one may use 
different approximation orders in each variable.
\end{remark}

\begin{remark}
The PICKAN model introduced here belongs to the class of \\ parametrized convex 
functions studied in \cite{schaller2025}.  
However, the architecture depicted in \ref{fig:PICKAN} differs from that 
of Schaller et al.: they consider a feedforward neural network whose weights and 
biases depend on $y$, with a nondecreasing convex activation function and 
nonnegative weights, while also taking $x$ as an input at each layer.
\end{remark}

We estimate the function 
\begin{equation}
  f(x,y)= |y+1| |x+2x^3| \mbox{ on } [-2,2]^2,
  \label{eq:fPartial}
\end{equation}
by minimizing the empirical version of
\begin{equation*}
    \E[ (f(X,Y) -\tilde f^\theta(X,Y))^2],
\end{equation*}
where $\tilde f^\theta$ is our neural network approximation parametrized by $\theta$ and $(X, Y) \sim U[-2,2]^2$.
As before, we use the ADAM optimizer with $200{,}000$ iterations and learning rate $10^{-3}$.
For the ICNN with ReLU activation, we test architectures with $2$, $3$, or $4$ layers, 
and hidden sizes in $\{10,20,40,80,160\}$.  
For each configuration, we record the average loss and standard deviation over 
$10$ independent runs.

As before, the best PICNN (partial input convex  neural network, \cite{amos2017input}) 
is used as the baseline.

Results are shown in \ref{tab:compICNN-ICKANPart} and indicate that the PICKAN achieves 
accuracy comparable to the best PICNN.  
The ratio of computational times between PICNN and P1–ICKAN remains 
consistent with the results reported in \ref{sec:ickan_num}.

\begin{table}[tbhp]
\centering
\caption{Testing partial convexity for function in \ref{eq:fPartial} : mean and standard deviation over 10 runs. {\it NBL} stands for the number of layers, {\it NBN} the number of neurons, {\it nbParam} the number of model parameters, while {\it Time} stands for the time in seconds for one hundred iterations of the stochastic gradient descent.}
    \label{tab:compICNN-ICKANPart}
    \begin{tabular}{|c|c|c|c|c|c|c|c|} \hline
method & NBL &  NBN & $P$  & Average & std   & nbParam & Time \\ \hline
Best PICNN  & 2 & 80 &  & 1.04E-03 & 7.37E-04 & 33126 & 0.28\\ \hline
PICKAN   & 2 & 20 & 20 & 4.98E-03 & 1.10E-02 & 18480& 0.77 \\ 
PICKAN   & 2 & 20 & 40 & 5.17E-03 & 4.87E-03 &36080 & 0.77 \\ 
PICKAN   & 2 & 40 & 20 & 2.57E-03 & 3.29E-03 & 70560& 0.78 \\ 
PICKAN   & 2 & 40 & 40 & 9.23E-04 & 5.66E-04 & 137760& 0.78\\ 
PICKAN   & 3 & 20 & 20 & 2.10E-03 & 3.55E-03  & 35280& 1.10\\ 
PICKAN   & 3 & 20 & 40 & 3.85E-03 & 3.69E-03 &68880 & 1.10 \\ 
PICKAN   & 3 & 40 & 20 & 3.75E-03 & 5.97E-03  &137760  & 1.11\\ 
PICKAN   & 3 & 40 & 40 & 9.34E-03 & 1.81E-02 &268960 &  1.11\\ \hline
PICKAN adapt  & 2 & 20 & 20 & 2.17E-03 & 2.58E-03 &20120 &  1.05\\ 
PICKAN adapt  & 2 & 20 & 40 & 2.67E-03 & 4.25E-03 &39360 & 1.05\\ 
PICKAN adapt  & 2 & 40 & 20 & 1.02E-03 & 1.03E-03 & 73800& 1.06\\ 
PICKAN adapt  & 2 & 40 & 40 & 1.27E-03 & 1.23E-03 &144240 & 1.06\\ 
PICKAN adapt  & 3 & 20 & 20 & 2.71E-03 & 3.98E-03 & 37720& 1.48\\ 
PICKAN adapt  & 3 & 20 & 40 & 1.10E-02 & 1.98E-02  &73760 & 1.48\\ 
PICKAN adapt  & 3 & 40 & 20 & 8.11E-03 & 1.65E-02 &142600 & 1.49 \\ 
PICKAN adapt  & 3 & 40 & 40 & 5.79E-03 & 5.07E-03 & 278640& 1.49\\ \hline
    \end{tabular}
\end{table}

\section{Application for optimal transport}
 \label{sec:optim_transport}
Neural networks that preserve convexity can be used to solve optimal transport problems; 
see for example \cite{makkuva2020optimal, korotin2021}, which employ ICNNs. 
In this section, we use ICKANs to solve optimal transport problems and compare 
their performance with ICNNs.

\subsection{Monge optimal transport and Brenier's theorem}
Let us consider two probability measures with support $\Omega \subset \R^d$, $\mu$ and $\nu$, such that $\int_{\Omega} \|x\|^2 d\mu(x)$, $\int_{\Omega} \|x\|^2 d\nu(x) < \infty$. Our goal is to estimate the optimal transport map between $X$ and $Y$, $T: \Omega \to \Omega$, which is a solution to the Monge problem~\cite{monge1781}
\begin{equation} \label{eq:monge}
\underset{T\#\mu = \nu}{\inf}  \int_{\Omega} \|x-T(x)\|^2d\mu(x),
\end{equation}
where $T\#P(\cdot) := P(T^{-1}(\cdot))$ denotes the pushforward of a measure $P$, 
and $\|\cdot\|$ is the Euclidean norm on $\mathbb{R}^d$.
Brenier's theorem \cite{brenier1991} guarantees existence and uniqueness of the solution 
to \ref{eq:monge}, provided that $\mu$ is absolutely continuous with respect to the 
Lebesgue measure, an assumption we make throughout.  
Furthermore, the optimal transport map can be written as the gradient of a convex 
function $f : \mathbb{R}^d \to \mathbb{R}$ that is differentiable almost everywhere.
Let $CVX(\mu)$ denote the set of convex functions in $L^{1}(\mu)$.  
Then $f$ solves
\begin{equation*}
\underset{\varphi \in CVX(\mu)}{\inf}  \int_{\Omega} \varphi d\mu+ \int_{\Omega} \varphi^{\star} d\nu,
\end{equation*}
where $\varphi^{\star}(y) = \underset{x \in \Omega}{\sup} \{ \langle x,y \rangle - f(x)\}$ for $y \in \Omega$ is the Legendre-Fenchel transform of $\varphi$, see~\cite[Theorem 1]{manole2024}. $f$ is called the Brenier's potential. In \cite{makkuva2020optimal}, it is  shown that when $\nu$ admits a density, 
there exists an optimal pair $(\varphi_0, \psi_0)$ solving
\begin{equation}
    \sup_{\begin{array}{c}
        \varphi \in CVX(\mu) \\
        \varphi^\star \in L^1(\nu)
    \end{array}}
     \inf_{\psi \in CVX(\nu)}
      - \int \varphi(x) d \mu(x)  - \int \Bigl( \langle y, \nabla \psi(y) \rangle - \varphi( \nabla \psi(y)) \Bigr) d \nu(y),
      \label{eq:forAlgo}
\end{equation}
where $\nabla \psi_0$ is the inverse of the optimal transport map. This formulation enables the use of convex neural networks to approximate 
both the optimal transport map and its inverse.

\subsection{Algorithm}
Given samples $X_1,\ldots,X_n \sim \mu$ and $Y_1,\ldots,Y_n \sim \nu$ both identically and independently distributed, the objective is to estimate the optimal transport map between a $\mu$ distribution and a $\nu$ distribution. We denote by $\hat \mu_n = \frac{1}{n} \sum_{i=1}^n \delta_{X_i}$ and $\hat \nu_n = \frac{1}{n} \sum_{i=1}^n \delta_{Y_i}$ the empirical distribution of $X$ and $Y$ respectively. 
In \ref{eq:forAlgo}, the potentials $\varphi$ and $\psi$ are parameterized by 
two distinct convex neural networks $\varphi_{\Theta}$ and $\psi_{\theta}$, 
which may be either ICKANs or ICNNs, with trainable parameters 
$\Theta,\theta \in \mathbb{R}^{l}$.
\\
The empirical counterpart of the minimax problem \ref{eq:forAlgo} becomes
\begin{equation*}
        \max_{\Theta} \min_{\theta} \frac{1}{n}  \sum_{i=1}^n\varphi_\Theta( \nabla \psi_\theta(Y_i)) - \langle Y_i, \nabla \psi_\theta(Y_i) \rangle - \varphi_\Theta(X_i).
\end{equation*}
We follow the classical minimax optimization scheme described in 
\ref{alg:MinMax}. The network parameter $\theta$ is minimized in an inner loop via several gradient‐descent steps while the other's $\Theta$ is maximized in an outer loop.
\begin{algorithm}[tbhp]
\caption{\bf Minimax algorithm for the transport problem \label{alg:MinMax}}
\begin{algorithmic}
 \Require Batch size $n$, number of outer iterations $I_{ext}$, number of inner  iteration $I_{int}$. 
  \For{$i=1, \dots I_{ext}$}
   \For{$j=1, \dots I_{int}$}
   \State  Sample $Y_1,\ldots,Y_n \sim \nu$ 
   \State  $\theta \leftarrow \text{argmin}_{\theta} \frac{1}{n}  \sum_{i=1}^n\varphi_\Theta( \nabla \psi_\theta(Y_i)) - \langle Y_i, \nabla \psi_\theta(Y_i) \rangle$ using Adam method
   \EndFor
     \State Sample $X_1,\ldots,X_n \sim \mu$ 
   \State  Sample $Y_1,\ldots,Y_n \sim \nu$ 
  \State $\Theta \leftarrow \text{argmax}_{\Theta} \frac{1}{n}  \sum_{i=1}^n\varphi_\Theta( \nabla \psi_\theta(Y_i)) - \langle Y_i, \nabla \psi_\theta(Y_i) \rangle - \varphi_\Theta(X_i)$ using Adam method
  \EndFor
  \end{algorithmic}
  \end{algorithm}
  Since the ICKAN networks are defined on a compact domain, and 
because $\nabla \psi(y)$ is not guaranteed to remain within this domain during optimization, 
the network $\psi_{\theta}$ is linearly extrapolated outside its domain of definition.

\subsection{Numerical results on synthetic data}
Since the optimal transport map is often unknown in closed form, we construct benchmark 
problems by choosing $\mu$ and a known transport map $T$ and defining $\nu = T \# \mu$.  
We assess the quality of the estimated map $\hat{T}$ on a validation dataset through the percentage of unexplained variance (UVP) introduced in \cite{korotin2021}:
\[
\mathrm{UVP}(\%) 
= 100 \times 
\frac{
\int_{\Omega} \|T^{\star}(x) - \hat{T}(x)\|^2\, d\hat{\mu}_n(x)
}{
\int_{\Omega} \|x\|^2\, d\hat{\nu}_n(x)
-
\Bigl\| \int_{\Omega} x\, d\hat{\nu}_n(x) \Bigr\|^2
}.
\]
A UVP of $100\%$ corresponds to the constant estimator 
$T^{C}(x) = \int_{\Omega} x\, d\hat{\nu}_n(x)$.

As in \cite{korotin2021}, we use as a benchmark the linear transport map
\[
\hat{T}^{L}(x)
= \hat{A}\,(x - \hat{m}_{1}) + \hat{m}_{2},
\]
where
\[
\hat{m}_{1} = \int_{\Omega} x\, d\hat{\mu}_n(x),
\qquad
\hat{m}_{2} = \int_{\Omega} x\, d\hat{\nu}_n(x),
\]
\[
\hat{\Sigma}_{1}
=
\int_{\Omega} x x^{\top}\, d\hat{\mu}_n(x)
-
\hat{m}_{1}\hat{m}_{1}^{\top},
\qquad
\hat{\Sigma}_{2}
=
\int_{\Omega} x x^{\top}\, d\hat{\nu}_n(x)
-
\hat{m}_{2}\hat{m}_{2}^{\top},
\]
and
\[
\hat{A}
=
\hat{\Sigma}_{1}^{-1/2}
\Bigl(
\hat{\Sigma}_{1}^{1/2}\,
\hat{\Sigma}_{2}\,
\hat{\Sigma}_{1}^{1/2}
\Bigr)^{1/2}
\hat{\Sigma}_{1}^{-1/2}.
\]

For the optimization, we use the same hyper-parameters as in \cite{korotin2021}:
\begin{itemize}
    \item $I_{\mathrm{ext}} = 50{,}000$ outer iterations, 
    \item $I_{\mathrm{int}} = 15$ inner iterations,
    \item batch size equal to $1024$,
    \item learning rate $0.001$,
    \item evaluation every $100$ iterations on a test set of size $4{,}096$, and we keep the network with the lowest error,
    \item validation dataset of size $2^{14}$.
\end{itemize}

For ICKANs, the domain is defined from the minimum and maximum values of a 
dataset of size $2^{14}$.  This framework is idealised, as in practice we by no means have this amount
of data, and we do not have access to the optimal map used to evaluate
the error on the test set. We also initialise the networks with $\theta_0$ minimising $|\nabla f_{\theta}(x) - x|^2$, so that the map is close to the identity at initialisation.

\begin{figure}[tbhp]
    \centering
    \begin{tabular}{ccc}
       \includegraphics[width=0.3\linewidth]{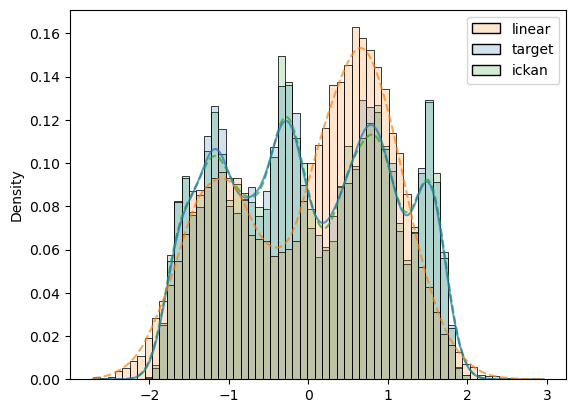}      &        \includegraphics[width=0.3\linewidth]{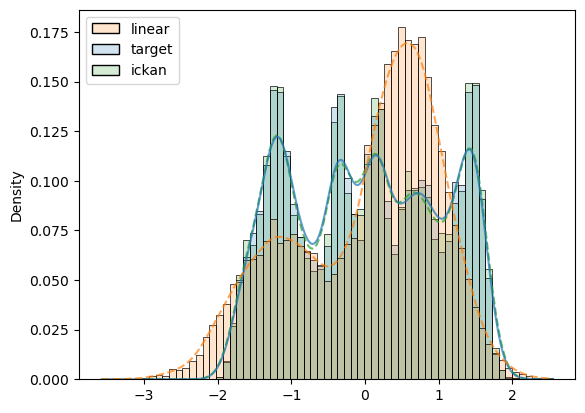} 
         &        \includegraphics[width=0.3\linewidth]{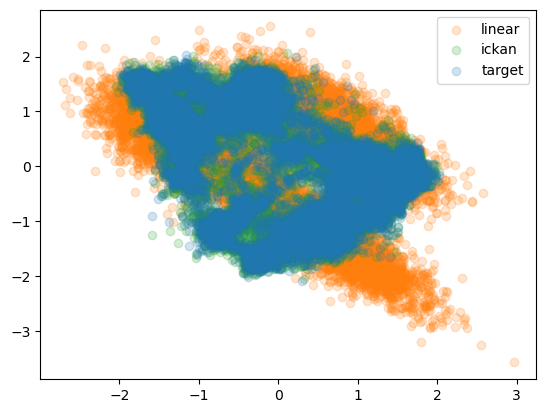} \\
         First marginal & Second marginal & Second versus first
    \end{tabular}
    \caption{Distribution of the true target  distribution as well as the one obtained by linear transport or Cubic-ICKAN transport with adapted mesh, $P=10$, and 64 and 32 neurons for the map in~\cite{korotin2021} and $d=2$. The first two figures include the empirical histogram as well as a Gaussian kernel density estimator with bandwidth selected using Scott's rule~\cite{scott2005} (solid line for target, dashed lines for linear and ICKAN transports).} 
    \label{fig:korotin2d}
\end{figure}

\medskip
\paragraph{\bf Transport map of~\cite{korotin2021}}
We consider the transport map introduced by Korotin et al.~\cite{korotin2021}.  
The authors study the optimal transport map $T_1$ between a mixture of 
three Gaussian distributions ($\mu$) and a mixture of ten Gaussians, and similarly 
$T_2$ between the same source mixture and another mixture of ten Gaussians.   They can then define the optimal transport map $\frac{1}{2}\left(T_1 + T_2\right)$. Since $T_1$ and $T_2$ are not explicitly known, they learn $T_1$ and $T_2$ with an ICNN, $\hat{T}_1$ and $\hat{T}_2$ respectively, that solves the optimal transport problem. The target distribution considered by Korotin et al.~\cite{korotin2021} and by us is then $\frac{1}{2}\left(\hat{T}_1 + \hat{T}_2\right) \# \mu$. We consider for the ICNN\footnote{We use the implementation at \url{https://github.com/iamalexkorotin/Wasserstein2Benchmark/}, which uses a CELU activation function.} 3 layers with $64$, $64$ and $32$ neurons (same parametrisation as in~\cite{korotin2021}) and for the ICKANs a network with 2 layers with $64$ and $32$ neurons or with $10$ and $5$ neurons. The results are given in \ref{tab:results_korotin} for $d \in \{2,4,4,8, 16, 32\}$. Our optimization framework and network parametrization correspond to the [MMv2] case in~\cite{korotin2021}. For the Cubic-ICKAN with adapted mesh and with 64 and 32 neurons, the errors are much smaller than for the linear map, but can be slightly larger than those obtained by the ICNN parametrization in some case, with similar orders of magnitude. As mentioned in~\cite{korotin2021}, the optimal transport map itself is learned with an ICNN, which can be advantageous for the ICNN parametrisation. Furthermore, we did not search for the optimal parametrisation of the Cubic-ICKAN (nor for the ICNN). The number of mesh points $P \in \{10, 20, 40\}$ does not have much influence on the results, as the use of an unadapted grid does. However, using a P1 mesh instead of a cubic one significantly degrades the results, but the network still outperforms the linear map. Using a smaller network of 10 and 5 neurons for the Cubic-ICKAN results in larger errors, and this difference increases with dimension. The obtained distributions are displayed in \ref{fig:korotin2d} for the case $d=2$ and the Cubic-ICKAN with $P=10$. 

\begin{table}[tbhp]
    \centering
    \caption{Percentage of unexplained variance UVP (\%) for the linear map, the map parametrized by the Cubic-ICKAN with adapted mesh and $P \in \{10, 20, 40\}$, the Cubic-ICKAN with non adapted mesh and $P=40$, the P1-ICKAN with adapted mesh and $P=40$ and the ICNN map when the true map is the one in~\cite{korotin2021}. For the Cubic-ICKAN, we consider networks with 2 layers and 64 and 32 neurons or 10 and 5 neurons, for the P1-ICKAN 2 layers with 64 and 32 neurons, while for the ICNN we consider 3 layers with 64, 64, and 32 neurons.} 
    \label{tab:results_korotin}
    \footnotesize

\begin{tabular}{llrrrrr}
\toprule
Method & Neurons / Dim & 2 & 4 & 8 & 16 & 32 \\
\midrule
Linear &  & 13.93 & 14.96 & 27.29 & 42.05 & 55.48 \\
\midrule
Cubic-ICKAN  P=10 adapt& 64 32 & 0.06 & 0.58 & 3.00 & 7.16 & 9.89 \\
Cubic-ICKAN P=20 adapt& 64 32 & 0.05 & 0.51 & 2.44 & 5.97 & 7.66 \\
Cubic-ICKAN P=40 adapt& 64 32 & 0.05 & 0.52 & 2.82 & 6.57 & 5.90 \\
\midrule
Cubic-ICKAN P=40  & 64 32 & 0.06 & 0.65 & 3.66 & 6.63 & 10.98 \\
\midrule
P1-ICKAN P=40 adapt & 64 32 & 1.01 & 6.62 & 19.99 & 19.90 & 27.95 \\
\midrule
Cubic-ICKAN P=40 adapt & 10 5 & 0.16 & 1.84 & 11.56 & 47.25 & 101.83 \\
\midrule
ICNN & 64 64 32 & 0.07 & 0.27 & 0.74 & 1.98 & 3.01 \\
\bottomrule
\end{tabular}

\end{table}

\begin{figure}[tbhp]
    \centering
    \begin{tabular}{ccc}
       \includegraphics[width=0.3\linewidth]{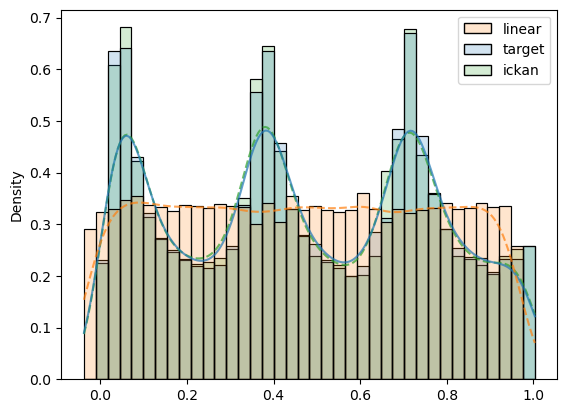}      &        \includegraphics[width=0.3\linewidth]{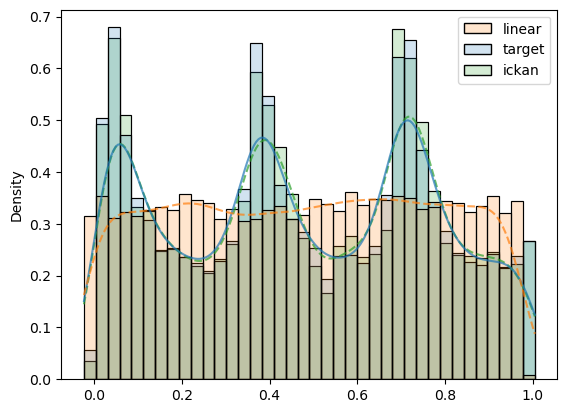} 
         &        \includegraphics[width=0.3\linewidth]{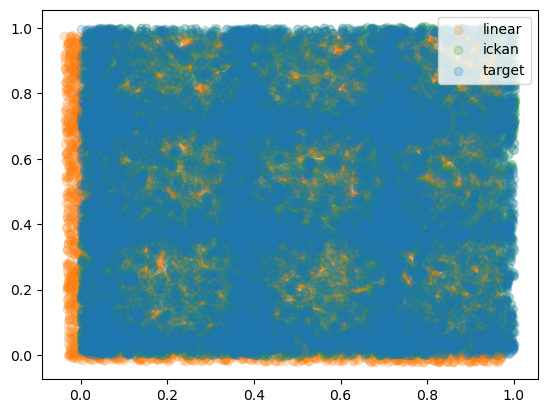} \\
         First marginal & Second marginal& Second versus first
    \end{tabular}
    \caption{Distribution of the true target distribution as well as the one obtained by linear transport or Cubic-ICKAN transport with adapted mesh and $P=10$, for the map $T(x) = (T_i(x_i))_{i=1,2}$ with $T_i(x_i) = x_i + \frac{1}{6 - \cos(2\pi x_i)}- 0.2$, $i=1,2$.  
    The first two figures include the empirical histogram as well as a Gaussian kernel density estimator with bandwidth selected using Scott's rule~\cite{scott2005} (solid line for target, dashed lines for linear and ICKAN transports).} 
    \label{fig:tensorized2d}
\end{figure}

\begin{figure}[tbhp]
    \centering
    \begin{tabular}{cc}
    \includegraphics[width=0.45\linewidth]{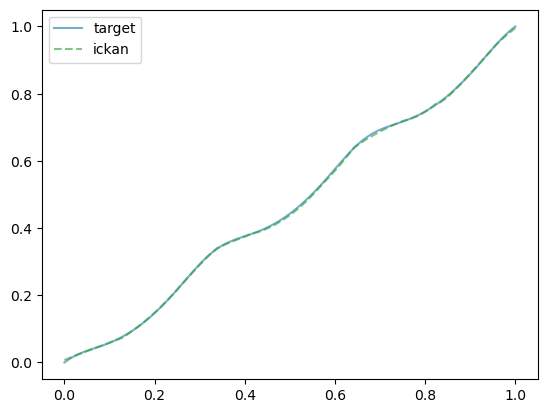}         &     \includegraphics[width=0.45\linewidth]{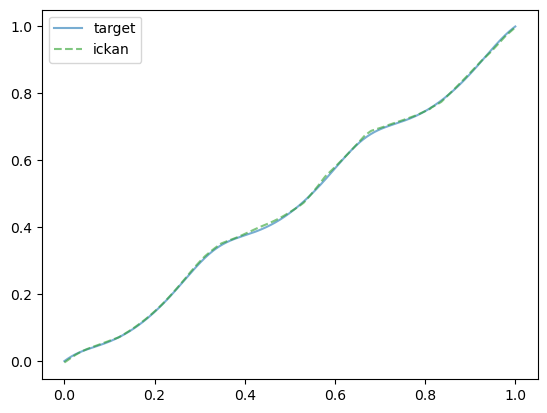}    \\
    \end{tabular}
    \caption{
    $x \mapsto T_1(x)$ (left) and $x \mapsto T_2(x)$ (right) for the map $T(x) = (T_i(x_i))_{i=1,2}$ with $T_i(x_i) = x_i + \frac{1}{6 - \cos(2\pi x_i)}- 0.2$, $i=1,2$ as well as the first component of $x \mapsto \hat{T}\bigl(\begin{pmatrix}x& 0.5\end{pmatrix}^{\top}\bigl)$ (left) and the second component of $x \mapsto \hat{T}\bigl(\begin{pmatrix}0.5&x\end{pmatrix}^\top\bigr)$ with $\hat{T}$ the estimated map parametrized by the Cubic-ICKAN with adapted mesh and $P=10$.}
    \label{fig:map_tensorized2d}
\end{figure}

\medskip
\paragraph{\bf Tensorized case~\cite{vacher2022}} We consider the example in Vacher et al.~\cite{vacher2022} where the transport map is defined by $T(x) = (T_i(x_i))_{i=1,\ldots,d}$ with $T_i(x_i) = x_i + \frac{1}{6 - \cos(2\pi x_i)}-0.2$, $i=1,\ldots,d$,\footnote{The example considered is the one from the code~\url{https://github.com/litlboy/OT-Model-Selection/blob/main/Synth-XP/Tensorised/sinkhorn.py} which differs from the one of the paper~\cite{vacher2022}.}  for $x = (x_i)_{i=1,\ldots,d}\in \left[0,1\right]^d$. $\mu$ is a uniform law on $\left[0,1\right]^d$. We consider a 3-layers ICNN with $64$, $64$, $32$ neurons and a 2-layers Cubic-ICKAN with $\max(2 d, 10)$ and $\max(d, 5)$ neurons, an adapted mesh and $P \in \{10, 20, 40\}$. The results are given in \ref{tab:results_tensorized} for $d \in \{1,2,4,8\}$. The ICNN performs very poorly for $d \in \{4,8\}$, giving errors of the same order of magnitude as the linear transport map. %, see also \ref{fig:tensorized2d_icnn}. 
The Cubic-ICKAN performs very well for all values of $P$. The obtained distributions are displayed in \ref{fig:tensorized2d} for the case $d=2$ case and the Cubic-ICKAN with $P=10$. We also display in \ref{fig:map_tensorized2d} the first component of $\hat{T}\bigl(\begin{pmatrix}x & 0.5\end{pmatrix}^{\top}\bigr)$ (0.5 is chosen arbitrarily, the first component of $x_2 \to \hat T\bigl(\begin{pmatrix}x &x_2\end{pmatrix}^{\top}\bigr)$ being constant and estimating $T_1(x)$), and the second component of $\hat{T}\bigl(\begin{pmatrix}
 0.5&x   
\end{pmatrix}^{\top}\bigr)$. We get similar figures for larger dimensions. The neural network reproduces the target distribution and the transport map very well. The better performance of ICKAN over ICNN  is probably due to the structure of the Brenier map, which is of the form $f(x) = \sum_{i=1}^d f_i(x_i)$ with $f_i(x_i) = \int_{0}^{x_i} T_i(s)ds$, $i=1,\ldots,d$: the functions in the Arnold-Kolmogorov representation~\ref{eq:KAN} then have a high degree of smoothness which can lead to a faster rate of convergence, see Proposition 2.2 and 2.3 in \cite{warin2024p1} in the non-convex case.

\begin{table}[tbhp]
    \centering
    \caption{Percentage of unexplained variance UVP (\%) for the linear map, the map parametrized by the Cubic-ICKAN with adapted mesh and $P \in \{10, 20, 40\}$, and the ICNN map and different dimensions when the true map is $T(x) = (T_i(x_i))_{i=1,\ldots,d}$ with $T_i(x_i) = x_i + \frac{1}{6 - \cos(2\pi x_i)}- 0.2$, $i=1,\ldots,d$. }
    \label{tab:results_tensorized}
\begin{tabular}{lrrrr}
\toprule
Method / Dim & 1 & 2 & 4 & 8 \\
\midrule
Linear & 0.49 & 0.54 & 0.52 & 0.53 \\
Cubic-ICKAN P=10 & 0.00 & 0.01 & 0.01 & 0.02 \\
Cubic-ICKAN P=20 & 0.00 & 0.02 & 0.02 & 0.02 \\
Cubic-ICKAN P=40 & 0.01 & 0.02 & 0.02 & 0.02 \\
ICNN & 0.04 & 0.05 & 0.34 & 0.51 \\
\bottomrule
\end{tabular}

\end{table}

%\begin{figure}[tbhp]
%    \centering
%    \begin{tabular}{ccc}
       %\includegraphics[width=0.3\linewidth]{tensorized2d_icnn/hist0.png}      &        \includegraphics[width=0.3\linewidth]{tensorized2d_icnn/hist1.png} 
        % &        \includegraphics[width=0.3\linewidth]{tensorized2d_icnn/scatter01.png} \\
         %First marginal & Second marginal & Second component versus first one
    %\end{tabular}
    %\caption{Distribution of the true target distribution as well as the one obtained by linear transport or ICNN transport, for the map $T(x) = (T_i(x))_{i=1,2}$ with $T_i(x) = x_i + \frac{1}{6 - \cos(2\pi x_i)}- 0.2$, $i=1,2$. The first two figures include the empirical histogram as well as a Gaussian kernel density estimator with bandwidth selected using Scott's rule~\cite{scott2005}.} 
  %  \label{fig:tensorized2d_icnn}
%\end{figure}

\paragraph{\bf{Product case}} 

We finally consider the map $T = \nabla f$  with $f$ the convex function defined on $\left[0,1\right]^d$ by
\begin{equation}
    \label{eq:f_product} f(x) = 3^{-d}\prod_{i=1}^d \left(x_i^2 + x_i + 1\right),
\end{equation}
and $\mu$ is the uniform law on $\left[0,1\right]^d$. The networks have the same architecture as in the second case (tensorized map). The results are given in \ref{tab:results_product}: ICKAN and ICNN give similar results and outperform the linear map benchmark (except for the case $d=1$ where the optimal transport map is linear).

\begin{table}[tbhp]   
\caption{Percentage of unexplained variance UVP (\%) for the linear map, the map parametrized by the Cubic-ICKAN with adapted mesh and $P=40$, and the ICNN map and different dimensions when the true map is $T(x) = \nabla f(x)$ with $f$ given in \ref{eq:f_product}.}
    \label{tab:results_product}
    \centering
\begin{tabular}{lrrrr}
\toprule
Method / Dim & 1 & 2 & 4 & 8 \\
\midrule
Linear & 0.00 & 6.90 & 16.43 & 33.57 \\
\midrule
Cubic-ICKAN P=40 & 0.01 & 3.15 & 2.33 & 2.85 \\
\midrule
ICNN & 0.00 & 2.77 & 1.81 & 1.30 \\
\bottomrule
\end{tabular}

\end{table}

\section{Conclusion}
Depending on the case, the new Kolmogorov-Arnold network may outperform the ICNN, though this depends heavily on the case. 
In particular, its performance varies according to the structure and regularity of the target function.
Therefore, this new network may provide an alternative to the ICNN, depending on the structure of the solution. One obvious example is when the solution is separable, as demonstrated in \ref{sec:optim_transport}. 

Additionally, Kolmogorov-Arnold networks are known to be more interpretable than multilayer perceptrons (MLPs) because they are compositions of one-dimensional functions that can easily be plotted, a feature appreciated by users needing to explain the obtained results. 

Similarly to the original spline KAN for which the computational cost was highly reduced using an implementation different from the original design\footnote{\url{https://github.com/Blealtan/efficient-KAN}}, more effective implementations of the ICKAN may be possible and are left to future research.

\newpage
\bibliographystyle{plain}
\bibliography{biblio}

\appendix

\
\section{One dimensional approximation with one layer}\label{appendix:0}
For the P1-ICKAN and the Cubic-ICKAN, 
we illustrate how a single layer can effectively approximate various one-dimensional convex functions using adaptation.
\subsection{Using P1-ICKAN}
\label{appendix:0_1}
\begin{figure}[tbhp]
     \centering
        \begin{minipage}[b]{0.49\linewidth}
    \centering
     \includegraphics[width=\textwidth]{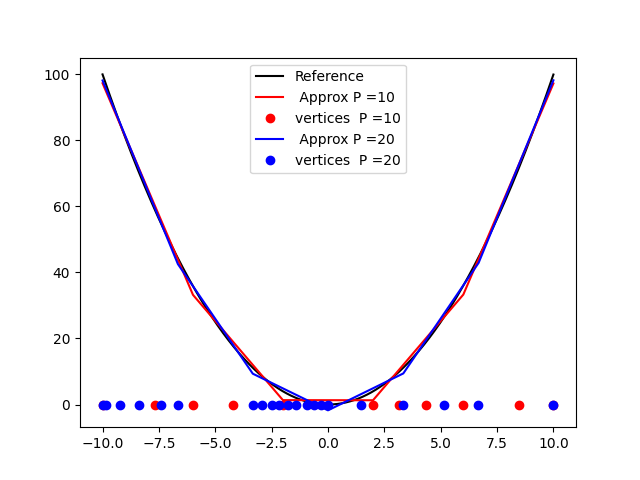}
    \caption*{$f_1$}
    \end{minipage}
           \begin{minipage}[b]{0.49\linewidth}
    \centering
     \includegraphics[width=\textwidth]{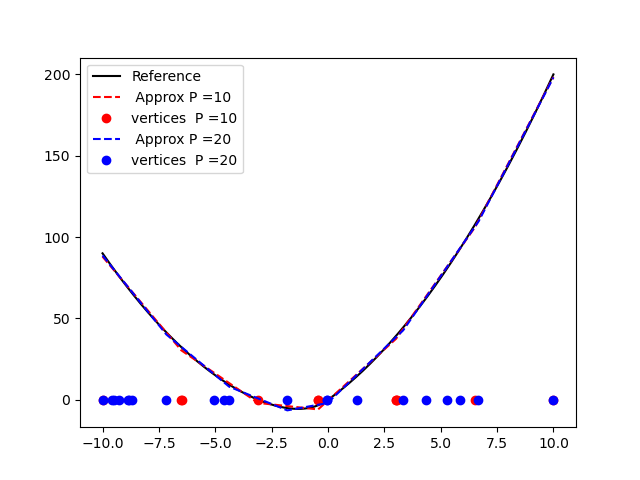}
    \caption*{$f_2$}
    \end{minipage}
           \begin{minipage}[b]{0.49\linewidth}
    \centering
     \includegraphics[width=\textwidth]{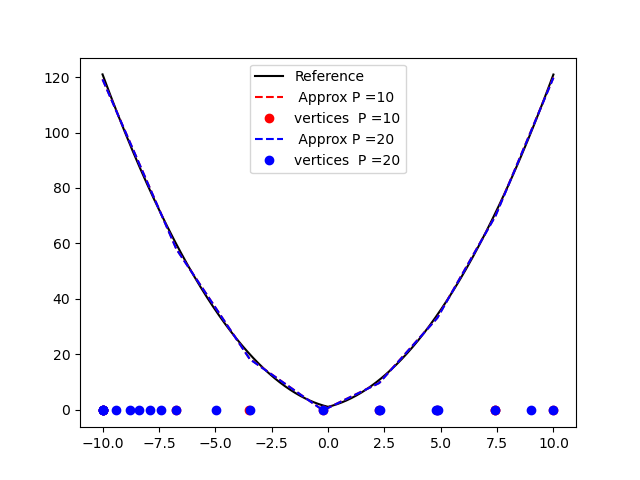}
    \caption*{$f_3$}
    \end{minipage}
           \begin{minipage}[b]{0.49\linewidth}
    \centering
     \includegraphics[width=\textwidth]{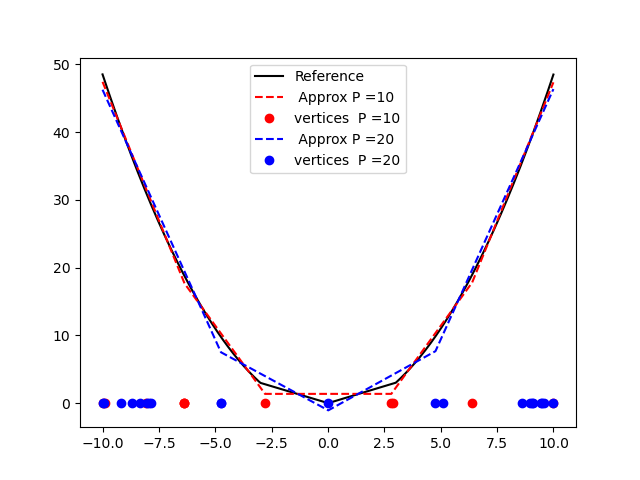}
    \caption*{$f_4$}
    \end{minipage}
     \caption{Piecewise linear approximation of a one dimensional function using  \ref{eq:P1}- \ref{eq:approxDer} with adaptation. \label{fig:approx1D}}
 \end{figure}

As an example, it is possible to minimize the mean squared error of a one dimensional convex function with its approximation \ref{eq:P1}-\ref{eq:approxDer} by training $b, \hat b, (d_i)_{i=1, \ldots, P-1}$, and $(e_i)_{i=1,\ldots,P}$. With $\theta$ the vector of parameters to be trained, our approximation $\tilde f^\theta$ of a function $f$ is parametrized by $\theta$ and 
we minimize
\[
    \E[ (f(X) -\tilde f^\theta(X))^2]
\]    
where $X$ is a random variable sampled uniformly on $[-10,10]$.
We give the results approximating the functions $f_i$, $i=1,\ldots,4$ on \ref{fig:approx1D} where
\begin{enumerate}
   \item $f_1(x)= x^2 $, 
   \item $f_2(x)= x^2 +  10 [(e^x-1) 1_{x<0} +x 1_{x\ge 0}]$ ,
     \item $f_3(x) = (|x|^2+1)^2$,
      \item $f_4(x) = |x| 1_{|x| \le 3} + \frac{x^2-3}{2}$.
\end{enumerate}
The plots are obtained adapting the grid with $P=10$ or $P=20$. We also plot the adapted vertices.

 \subsection{Cubic-ICKAN}
 \label{appendix:0_2}
\begin{figure}[tbhp]
     \centering
        \begin{minipage}[b]{0.49\linewidth}
    \centering
     \includegraphics[width=\textwidth]{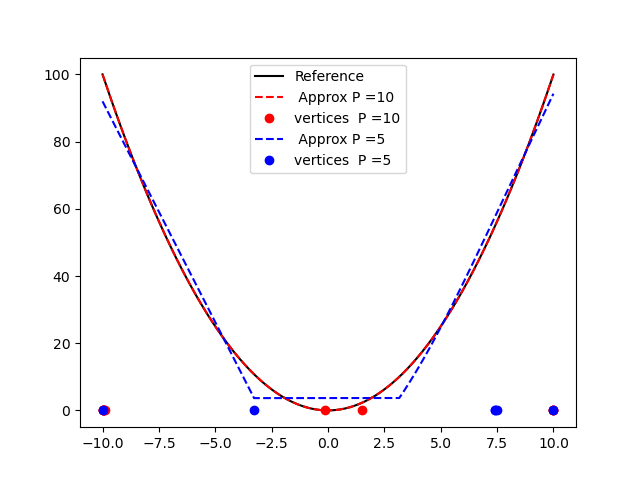}
    \caption*{$f_1$}
    \end{minipage}
           \begin{minipage}[b]{0.49\linewidth}
    \centering
     \includegraphics[width=\textwidth]{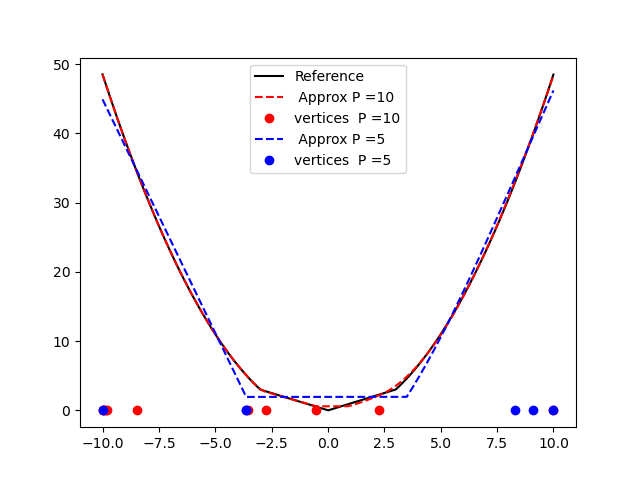}
    \caption*{$f_4$}
    \end{minipage}
     \caption{Cubic spline approximation of a one dimensional function using  \ref{eq:cubic1D}- \ref{eq:cubic1DExp} with adaptation. \label{fig:approx1DCubic}}
 \end{figure}

 As for the piecewise linear approximation, we provide on \ref{fig:approx1DCubic} the vertices and an estimation obtained by adapting the grid with $P=5$ or $P=10$.
%Compared with the P1 approximation, the cubic formulation better captures smooth variations of the target function, although at the cost of additional trainable parameters and slightly higher computational complexity.

\section{Proof of \ref{theo:universelAdapt} and \ref{theo:universelNoAdapt}}
\label{appendix:1}

\subsection{Proof of \ref{theo:universelAdapt}}
\label{appendix:proof1}

The proof is constructive and is based on the proof of theorem 1 in \cite{chen2018optimal}. One wants to approximate a Lipschitz convex function on $[0,1]^n$ denoted $f$ by a P1-ICKAN network with adaptation. First any Lipschitz convex function on $[0,1]^n$ can be approximated within $\epsilon$ by the maximum of a finite set of affine functions (Lemma 1 in \cite{chen2018optimal} or Proposition 1 in \cite{warin2023groupmax}), that is for any $\epsilon > 0$, there exists $N \geq 1$ and a function $f_N$ such that $|f(x)-f_N(x)| < \epsilon$ for any $x \in \left[0,1\right]^n$, with a representation   
\begin{equation} \label{eq:convex_affine}
f_N(x) := \,\max(\alpha_1^\top x + \beta_1,  \ldots , \alpha_N^\top x+ \beta_N),\;    \alpha_i \in \R^n,\,\beta_i \in \R,\,i=1,\ldots,N.
\end{equation}
As a consequence, the proof of \ref{theo:universelAdapt} boils down to show that the P1-ICKAN network with adaptation can represent any maximum of a finite set of affine functions of type \ref{eq:convex_affine} for any $N \geq 1$.

\medskip
Following \cite{chen2018optimal}, we start by the case $N=2$ and begin to show that the linear KAN can represent the maximum of two affine functions, that is any function of the form $f_2$. First, we have for $x \in [0,1]^n$, $\alpha_i \in \R^n$, $\beta_i \in \R$, $i=1,2$,

\begin{equation}\label{eq:f2_convex}
    f_2(x) = \max((\alpha_1^\top - \alpha_2^\top)x + \beta_1-\beta_2, 0) +  \alpha_2^\top x+ \beta_2. \\
\end{equation}
We consider $P=2$ and a $L=3$-layers neural network to match \ref{eq:f2_convex} with $n_1=3$ neurons for the first layer and $n_2=2$ neurons for the second layer. For the first layer, we consider the lattice defined by $\hat{x}_{0,j,0} = 0$, $\hat{x}_{0,j,1} = \frac{1}{2}$, and $\hat{x}_{0,j,2} = 1$, $1 \le j \le n$, and we define in \ref{eq:firstLPart} the parameters

\begin{equation}
\begin{split}
  b_{0,1,1}=& \, \beta_1 , \\
  b_{0,1,j} = & \, 0  \mbox{ for } 1 < j \le n , \\
  \hat b_{0,1,j}= &\,   \alpha_{1,j} \mbox{ for }  1 \le j \le n,  \\
    d_{0,1,j,1}= &\, 0  \mbox{ for }  1 \le j \le n,\\
  b_{0,2,1} = & \, -\beta_2,  \\
  b_{0,2,j} = & \, 0  \mbox{ for }  1 < j \le n ,  \\
  \hat b_{0,2,j} = &\,  -\alpha_{2,j}  \mbox{ for }  1 \le j \le n,\\
      d_{0,2,j,1}= &\, 0  \mbox{ for }  1 \le j \le n,\\
  b_{0,3,1} = & \, \beta_2,  \\
  b_{0,3,j} = & \, 0  \mbox{ for }  1 < j \le n ,  \\
  \hat b_{0,3,j} = &\,  \alpha_{2,j}  \mbox{ for }  1 \le j \le n,\\
      d_{0,3,j,1}= &\, 0  \mbox{ for }  1 \le j \le n.
  \end{split}
  \label{eq:disProof²}
\end{equation}
One checks that the output of the first layer is given by
\begin{align*}
    \hat \kappa^0_{n,n_1}(x,[0,1]^n)_1=&\, \alpha_1^\top x + \beta_1, \\
       \hat \kappa^0_{n,n_1}(x,[0,1]^n)_2 =&\, -\alpha_2^\top x - \beta_2,\\
           \hat \kappa^0_{n,n_1}(x,[0,1]^n)_3 =&\, \alpha_2^\top x + \beta_2.\\
\end{align*}

\medskip
For the second layer, we have $n_2 = 2$ neurons. Let $I_1 = \displaystyle{\prod_{j=1}^3} [\hat x_{1,j,0},\hat x_{1,j,2}]$ denote the image of $[0,1]^n$ by the first layer and consider for each $j=1,2,3$ any $\hat x_{1,j,1} \in ]\hat x_{1,j,0},\hat x_{1,j,2}[$ to define the grid $\mathcal{G}^1$. For the first neuron, we consider the parameters 
\begin{equation*}
    \begin{split}
& b_{1,1,j} = \hat x_{1,j,0},\,j=1,2,\\
& \hat b_{1,1,j}= 1,\,j=1,2,\\
& d_{1,1,j,1} = 0,\,j=1,2, \\
&b_{1,1,3}= 0, \\
&\hat b_{1,1,3}= 0,\\
&d_{1,1,3,1} = 0
\end{split}
\end{equation*}
in order to get 
\[\hat{\kappa}^1_{n_1,n_2}(x, \mathcal{G}_1)_1 = x_1 + x_2.\]
In the same way, for the second neuron, we consider the parameters 
\begin{equation*}
    \begin{split}
& b_{1,2,j} = 0,\,j=1,2,\\
& \hat b_{1,2,j}= 0,\,j=1,2,\\
& d_{1,2,j,1} = 0,\,j=1,2, \\
&b_{1,2,3}= \hat{x}_{1,3,0}, \\
&\hat b_{1,2,3}= 1,\\
&d_{1,2,3,1} = 0
\end{split}
\end{equation*}
to get 
\[\hat{\kappa}^1_{n_1,n_2}(x, \mathcal{G}_1)_2 = x_3.\]
At this stage, we then have as output of the two first layers
\begin{align*}
\hat \kappa^1_{n_1,n_2} \circ \kappa^0_{n,n_1}(x,[0,1]^n)_1&= (\alpha_1^\top - \alpha_2^\top)x + \beta_1-\beta_2,\\
\hat \kappa^1_{n_1,n_2} \circ \kappa^0_{n,n_2}(x,[0,1]^n)_2&=\alpha_2^{\top}x + \beta_2.
\end{align*}

\medskip
Let $I_2 =\displaystyle{\prod_{j=1}^2} [\hat x_{2,j,0},\hat x_{2,j,2}]$ denote the image of $\mathcal{G}_1$ by the second layer. For the third layer, we first consider the case where the non linearity is active for the first component of the second layer output $\hat \kappa^1_{n_1,n_2} \circ \kappa^0_{n,n_1}(x,[0,1]^n)_1$, that is $0 \in \left]\hat x_{2,1,0},\hat x_{2,1,2}\right[$ (only depending on the function that we try to equalize, $f_2$). In this case, we take $\hat x_{2,1,1} =0 \in \left]\hat x_{2,1,0},\hat x_{2,1,2}\right[$, and any $\hat x_{2,2,1} \in ]\hat x_{2,2,0},\hat x_{2,2,2}[$ to construct $\mathcal{G}_2$, and define
%(meaning that $0$ is  in the interior of the image $I = \displaystyle{\prod_{i=1}^2} [\hat x_{1,i,0},\hat x_{1,i,2}] $ of $[0,1]^n$ by the first layer),  take $\hat x_{1,1,1} =0$  and  define in \ref{eq:secLayer}
\begin{equation}
\begin{split}
\label{eq:nn_second_layer}
&b_{2,1,1}= 0, \\
&\hat b_{2,1,1} = 0,\\
&d_{2,1,1,1} =  1, \\
&b_{2,1,2}= \hat x_{2,2,0}, \\
&\hat b_{2,1,2}= 1,\\
&d_{2,1,2,1} = 0.
\end{split}
\end{equation}
The three first lines in~\ref{eq:nn_second_layer} define the ReLU with an active non linearity, while the three last define the identity function.
We thus get the desired result
\begin{align*}
\hat \kappa^1_{n_2,1} \circ \kappa^1_{n_1,n_2} \circ \kappa^0_{n,n_1}(x,[0,1]^n)_1=&\, \max((\alpha_1^\top - \alpha_2^\top)x + \beta_1-\beta_2, 0) +  \alpha_2^\top x+ \beta_2
\end{align*}
equal to $f_2(x)$ from~\ref{eq:f2_convex}.
Now consider the case when the non linearity is not active. It is enough to take for the lattice any $\hat{x}_{2,1,1} \in ]\hat{x}_{2,1,0}, \hat{x}_{2,1,2}[$ and to replace $b_{2,1,1}$, $\hat b_{2,1,1}$, and $d_{2,1,1,1}$ in \ref{eq:nn_second_layer} by
\begin{equation*}
    \begin{split}
& b_{2,1,1} = \hat x_{2,1,0},\\
& \hat b_{2,1,1}= 1, \\
& d_{2,1,1,1} = 0
\end{split}
\end{equation*}
to obtain the identity function.

\medskip

Similarly as in \cite{chen2018optimal}, one can extend iteratively the procedure for a given $N$ using
%\begin{equation} \label{eq:recursive}
%        f_N(x) =  \max( f_{N-1}(x),  \alpha_N^\top x+ \beta_N).
%\end{equation}
\begin{equation} \label{eq:recursive}
        f_N(x) =  \max( f_{N-1}(x) -(\alpha_N^\top x+ \beta_N),0)+  \alpha_N^\top x+ \beta_N).
\end{equation}
If one has constructed $f_{N-1}(x)$, add 
\begin{itemize}
    \item Two neurons at the end of the first layer, with outputs $-\alpha_N^{\top}x - \beta_N$ and $\alpha_N^{\top} x + \beta_N$, in the same way as the last two neurons in the first layer in the case $N=2$,
    \item Two neurons at the end of each existing layer, that output exactly the outputs of the two last neurons of the previous layer, in the same way that the last neuron in the second layer in the case $N=2$, and we get $-\alpha_N^{\top}x - \beta_N$ and $\alpha_N^{\top} x + \beta_N$ for the outputs of the two last neurons of the last layer.
\end{itemize}
From this, we get a third dimensional output corresponding to $f_{N-1}(x)$, $-\alpha_N^{\top}x - \beta_N$ and $\alpha_N^{\top} x + \beta_N$. The situation is exactly the same as for the case $N=2$, after that the first layer has been constructed. It remains to add the two layers similar to the two last layers in the case $N=2$ to get the desired output~\ref{eq:recursive}. 
%used in the construction of the neural network in the case $N=2$ $\hat{\kappa}^1_{n_1,n_2}$ and $\hat{\kappa}^2_{n_1,n_2}$, to get the desired output~\ref{eq:recursive}. 

\begin{remark}
The first layer outputs $-\alpha_N^{\top} x - \beta_N$ in order to avoid to have negative values of $\hat{b}_{l,k,j}$ in the other layers which, contrarily to the first layer, consider $\max(\hat{b}_{l,k,j},0)$ and not $\hat{b}_{l,k,j}$. This is similar to the duplication trick in the proof of theorem 1 in~\cite{chen2018optimal}.
\end{remark}

\subsection{Proof of \ref{theo:universelNoAdapt}}
\label{appendix:proof2}
The idea of the proof is the same as in the adapted case. 
The main difference is that, without mesh adaptation, we must control the approximation error by increasing the number of mesh points $P$.

For simplicity, we allow here the number of mesh points $P+1$ to vary from one neuron to another. This has no impact because a neuron with a mesh size of $M + 1$ is a special case of a neuron with a mesh size of $P + 1$ if $M \leq P$. 

We first consider the case $N=2$ as in the adapted proof.  
The construction of the first two layers is identical.  
The only difference appears in the third layer, when the nonlinearity is active, a priori
\[
0 \notin ]\hat x_{2,1,0}, \hat x_{2,1,2}[
\]
and the ReLU function must be approximated even though. For the first neuron of this layer, given $P$, there exists $k$ such that
\[
|\hat x_{2,1,k}| \le \frac{\hat x_{2,1,P}-\hat x_{2,1,0}}{P}.
\]
We then choose parameters
\begin{equation*}
\begin{split}
\label{eq:nn_second_layer_noAdapt_0}
&b_{2,1,1}= 0, \\
&\hat b_{2,1,1} = 0,\\
&d_{2,1,1,i} =  0, \mbox{ for } i \neq k \\
& d_{2,1,1,k}=1
\end{split}
\end{equation*}
and we keep $P=2$ for the second neuron and parameters
\begin{equation*}
\begin{split}
\label{eq:nn_second_layer_noAdapt_1}
&b_{2,1,2}= \hat x_{2,2,0}, \\
&\hat b_{2,1,2}= 1,\\
&d_{2,1,2,1} = 0.
\end{split}
\end{equation*}
The output of the neural network is then
\begin{align*}
\kappa_2(x) &:=\hat \kappa^1_{n_2,1} \circ \kappa^1_{n_1,n_2} \circ \kappa^0_{n,n_1}(x,[0,1]^n)_1\\
&= \max((\alpha_1^\top - \alpha_2^\top)x + \beta_1-\beta_2 -\hat x_{2,1,k}, 0) +  \alpha_2^\top x+ \beta_2
\end{align*}
so that
\begin{equation*}
    | f_2(x) - \kappa_2(x) | \le |\hat x_{2,1,k}| \le \frac{C_2}{P}.
\end{equation*}
The previous result can be extended to the general case $N \geq 2$ so that our network representation $\kappa_N$ is such that
\begin{equation*}
    | f_N(x) - \kappa_N(x) | \le \frac{C_N}{P}
\end{equation*}
and
\begin{equation*}
    |f(x) - \kappa_N(x) | \le \frac{C_N}{P} +\epsilon.
\end{equation*}
Therefore, setting $P$ equal to the integer part of $\frac{C_N}{\epsilon} + 1$ provides a $2\epsilon$ approximation of $f$.

\end{document}